\documentclass[10pt,journal,compsoc]{IEEEtran}

\usepackage{times}
\usepackage{epsfig}
\usepackage{graphicx}
\usepackage{amsmath}
\usepackage{amssymb}

\usepackage{multirow}
\usepackage{textcomp}
\usepackage{subfiles}
\usepackage{adjustbox}
\usepackage{siunitx}
\usepackage[inline, shortlabels]{enumitem}
\usepackage{xspace}
\usepackage[hidelinks]{hyperref}

\newcommand*{\eg}{e.g.\@\xspace}

\newcommand{\etal}{\textit{et al}.}
\makeatletter
\newcommand*{\etc}{%
    \@ifnextchar{.}%
        {etc}%
        {etc.\@\xspace}%
}
\makeatother

\DeclareMathOperator*{\argmin}{arg\,min}
\newcommand{\ma}[1]{\mathrm{#1}}

\newcommand{\norm}[1]{\left\lVert#1\right\rVert}
\newcommand{\st}{\text{subject to}}

\newcommand{\round}[2]{\num[round-mode=places,round-precision=#1]{#2}}

\newcommand{\lcoef}{\boldsymbol{\beta}}
\newcommand{\lcoefi}{\beta_i}
\newcommand{\lcoefx}[1]{\beta_#1}
\newcommand{\wtwo}{\ma{W}}
\newcommand{\wtwoi}{\ma{W_i}}
\newcommand{\wtwox}[1]{\ma{W_#1}}
\newcommand{\reW}{\ma{W^\prime}}
\newcommand{\reY}{\ma{Y}}
\newcommand{\reX}{\ma{X}}
\newcommand{\reXi}{\ma{X_i}}
\newcommand{\reXx}[1]{\ma{X_#1}}
\newcommand{\lalpha}{\lambda}
\newcommand{\rank}{c}
\newcommand{\out}{n}
\newcommand{\kh}{k_h}
\newcommand{\kw}{k_w}
\newcommand{\samp}{N}
\newcommand{\rankp}{c^\prime}

\newcommand{\wxi}{\ma{Z_i}}
\newcommand{\cx}{\ma{X^\prime}}

\newcommand{\x}[1]{$#1\times$}

\newcommand{\las}{LASSO regression}
\newcommand{\conv}{convolutional layer}
\newcommand{\featch}{feature map width}

\newcommand{\ratio}{speed-up ratio}

\newcommand{\firstk}{\textit{first k}}
\newcommand{\prune}{\textit{max response}}

\newcommand{\filterwise}{\textit{filter-wise pruning}}
\newcommand{\Filterwise}{\textit{Filter-wise pruning}}

\newcommand{\sampling}{\textit{feature map sampling}}

\newcommand{\filterpruning}{Filter pruning}
\newcommand{\tensordecom}{tensor factorization}
\newcommand{\Tensordecom}{Tensor factorization}
\newcommand{\TensorDecom}{Tensor Factorization}
\newcommand{\channelpruning}{channel pruning}
\newcommand{\Channelpruning}{Channel pruning}

\newcommand{\structured}{structured simplification}
\newcommand{\Structured}{Structured simplification}
\newcommand{\implementation}{optimized implementation}
\newcommand{\Implementation}{Optimized implementation}
\newcommand{\sparseconnect}{sparse connection}
\newcommand{\Sparseconnect}{Sparse connection}

\newcommand{\RNP}{RNP~\cite{lin2017runtime}}
\newcommand{\thinet}{ThiNet~\cite{luo2017thinet}}
\newcommand{\spp}{SPP~\cite{wang2017structured}}

    \newcommand{\tworow}[1]{\begin{tabular}[c]{@{}c@{}}#1\end{tabular}}
    \newcommand{\jadercite}{Jaderberg \etal~\cite{jaderberg2014speeding}}
    \newcommand{\jader}{Jaderberg \etal~\cite{jaderberg2014speeding} (\cite{Zhang2015}'s impl.)}
    \newcommand{\asym}{Asym.~\cite{Zhang2015}}
    \newcommand{\asymd}{Asym. 3D~\cite{Zhang2015}}
    \newcommand{\asymdft}{Asym. 3D (fine-tuned)~\cite{Zhang2015}}
    \newcommand{\prli}{\filterpruning~\cite{Li2016}}
    \newcommand{\pr}{\filterpruning~\cite{Li2016} (our impl.)}
    \newcommand{\prft}{\filterpruning~\cite{Li2016} (fine-tuned, our impl.)}
    \newcommand{\prfttwo}{\tworow{\filterpruning~\cite{Li2016}\\ (fine-tuned, our impl.)}}
    \newcommand{\sgd}{\textit{SGD}}

\newcommand{\origvgg}{89.9}
\newcommand{\vggtworaw}{2.7}
\newcommand{\vggfour}{1.0}
\newcommand{\vggfouracc}{11.1} 
\newcommand{\vggfourraw}{7.9}
\newcommand{\vggfourrawacc}{18.0} 
\newcommand{\vggfive}{1.7}
\newcommand{\vggfiveraw}{22.0}
\newcommand{\vggscratchacc}{11.9} 
\newcommand{\vggscratcherr}{1.8}
\newcommand{\vggscratchuniacc}{12.5}
\newcommand{\vggscratchunierr}{2.4}

\newcommand{\vggc}{1.3}
\newcommand{\vggcft}{0.3}
\newcommand{\vggcfour}{0.7} 
\newcommand{\vggcftfour}{0.0}

\newcommand{\resorig}{92.2} 
\newcommand{\resmb}{4.0} 

\newcommand{\resft}{1.4}

\newcommand{\xceptionfifty}{Xception-50}
\newcommand{\xceptionpr}{4.3}
\newcommand{\xceptionorig}{92.8}
\newcommand{\xceptioncr}{2.9}
\newcommand{\xceptionft}{1.0}

\ifCLASSOPTIONcompsoc
  \usepackage[nocompress]{cite}
\else
  \usepackage{cite}
\fi

\ifCLASSINFOpdf
\else
\fi

\begin{document}

\title{Pruning Very Deep Neural Network Channels for Efficient Inference}

\author{Yihui~He
\IEEEcompsocitemizethanks{\IEEEcompsocthanksitem  \href{https://yihui.dev/channel-pruning-for-accelerating-very-deep-neural-networks}{yihui.dev/channel-pruning-for-accelerating-very-deep-neural-networks}
}
}

\markboth{IEEE TRANSACTIONS ON PATTERN ANALYSIS AND MACHINE INTELLIGENCE}
{He \MakeLowercase{\textit{et al.}}: Pruning Very Deep Neural Network Channels for Efficient Inference}

\IEEEtitleabstractindextext{%
\begin{abstract}
    In this paper, we introduce a new channel pruning method to accelerate very deep convolutional neural networks. Given a trained CNN model, we propose an iterative two-step algorithm to effectively prune each layer, by a \las\ based channel selection and least square reconstruction. We further generalize this algorithm to multi-layer and multi-branch cases. Our method reduces the accumulated error and enhances the compatibility with various architectures. Our pruned VGG-16 achieves the state-of-the-art results by \x{5} speed-up along with only 0.3\% increase of error. More importantly, our method is able to accelerate modern networks like ResNet, Xception and suffers only \resft\%, \xceptionft\% accuracy loss under \x{2} speed-up respectively, which is significant. Our code has been made publicly available.
\end{abstract}

}

\maketitle

\IEEEdisplaynontitleabstractindextext

\IEEEpeerreviewmaketitle

\ifCLASSOPTIONcompsoc
\IEEEraisesectionheading{\section{Introduction}\label{sec:introduction}}
\else
\section{Introduction}
\label{sec:introduction}
\fi

    \IEEEPARstart{R}{ecently}, convolutional neural networks (CNNs) are widely used on embed systems like smartphones and self-driving cars. 
    The general trend since the past few years has been that the networks have grown deeper, 
    with an overall increase in the number of parameters and convolution operations. 
    \textit{Efficient inference} is becoming more and more crucial for CNNs~\cite{szegedy2015rethinking}.
   CNN acceleration works fall into three categories: \implementation\ (\eg, FFT~\cite{vasilache2014fast}), 
    quantization (\eg, BinaryNet~\cite{courbariaux2016binarynet}), 
    and \structured\ that convert a CNN into compact one~\cite{jaderberg2014speeding}.
    This work focuses on the last one since it directly deals with the over-parameterization of CNNs.
    
\begin{figure}
    \begin{center}
        \includegraphics[width=\linewidth]{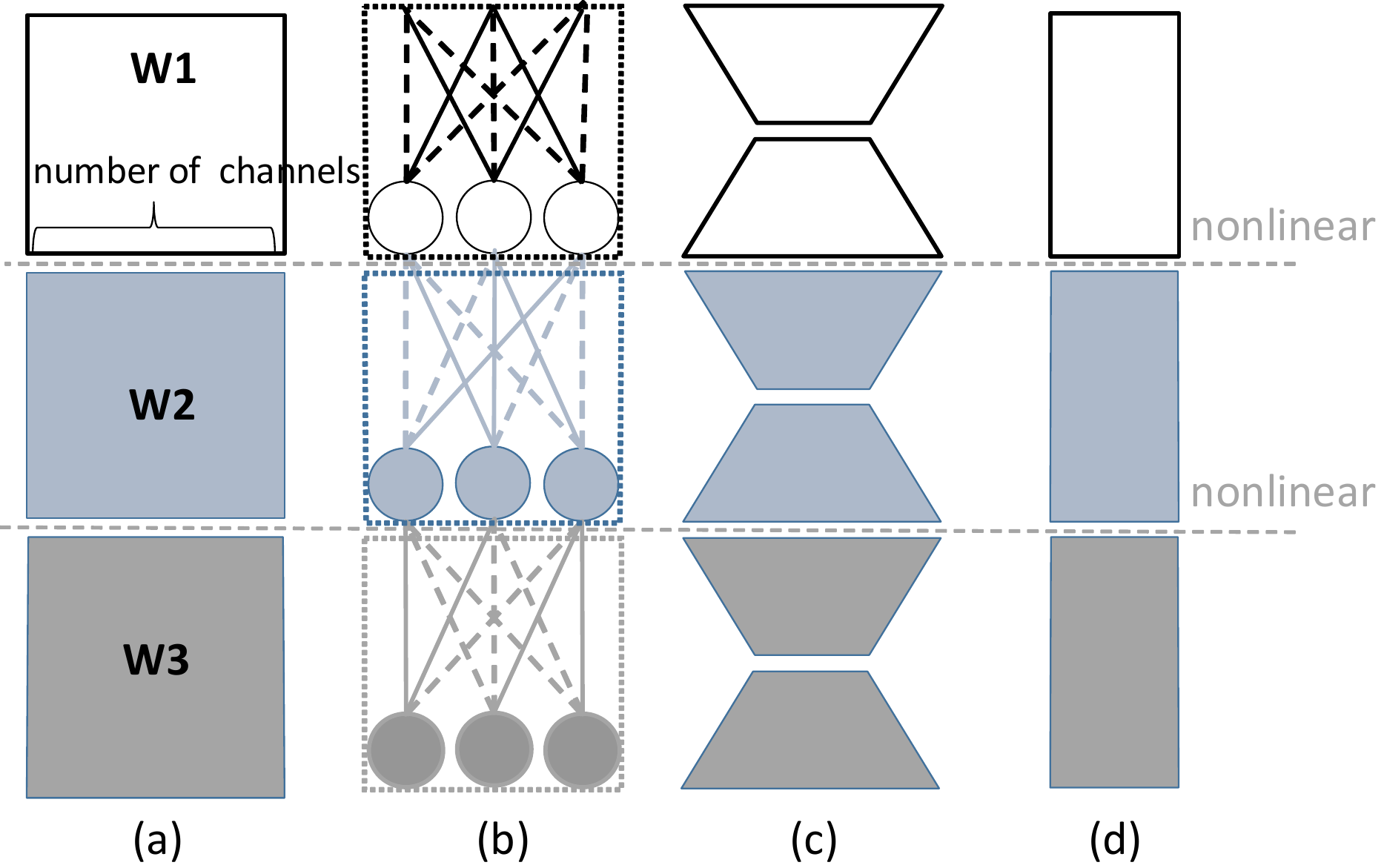}
    \end{center}
    \caption{\Structured\ methods that accelerate CNNs: (a) a network with 3 conv layers. (b) \sparseconnect\ deactivates some connections between channels. (c) \tensordecom\ factorizes a \conv\ into several pieces. (d) \channelpruning\ reduces number of channels in each layer (focus of this paper).}
    \label{fig:struct}
    \end{figure}

    \Structured\ mainly involves:
    \tensordecom~\cite{jaderberg2014speeding},
    \sparseconnect~\cite{han2015learning},
    and \channelpruning~\cite{wen2016learning}.
    \Tensordecom\ factorizes a \conv\ into several efficient ones (Fig.~\ref{fig:struct}(c)).
    However, \featch\ (number of channels) could not be reduced,
    which makes it difficult to decompose $1\times1$ \conv s favored by modern networks (\eg, GoogleNet~\cite{szegedy2015going}, ResNet~\cite{He2015},  Xception~\cite{chollet2016xception}). This type of method also introduces extra computation overhead.
    \Sparseconnect\ deactivates connections between neurons or channels (Fig.~\ref{fig:struct}(b)).
    Though it is able to achieve high theoretical \ratio,
    the sparse \conv s have an "irregular" shape which is not implementation-friendly.
    In contrast, \channelpruning\ directly reduces \featch, 
    which shrinks a network into thinner one, 
    as shown in Fig.~\ref{fig:struct}(d).
    It is efficient on both CPU and GPU because no special implementation is required.

   Channel pruning is simple but challenging because removing channels in one layer might dramatically change the input of the following layer. Recently, \textit{training-based} \channelpruning\ works~\cite{Alvarez2016,wen2016learning} have focused on imposing the sparse constraint on weights during training, which could adaptively determine hyper-parameters.
    However, training from scratch is very costly, and results for very deep CNNs on ImageNet have rarely been reported.
    \textit{Inference-time} attempts~\cite{Li2016,anwar2016compact} have focused on analysis of the importance of individual weight. The reported \ratio\ is very limited. 
    
        This work is initially inspired by Net2widerNet~\cite{Chen2015}, which could easily explore new wider networks specification of the same architecture.
    It makes a convolutional layer wider by creating several copies of itself, calling each copy so that the output feature map is unchanged.
    It's nature to ask: 
    \textit{Is it possible to convert a net to thinner net of the same architecture without losing accuracy?}
    If we regard each feature map as a vector, then all feature maps form a vector space.
    The inverse operation of Net2widerNet discussed above is to find a set of base feature vectors and use them to represent other feature vectors, in order to reconstruct the original feature map space. 

\begin{figure*}
        \begin{center}
        \includegraphics[width=0.7\linewidth]{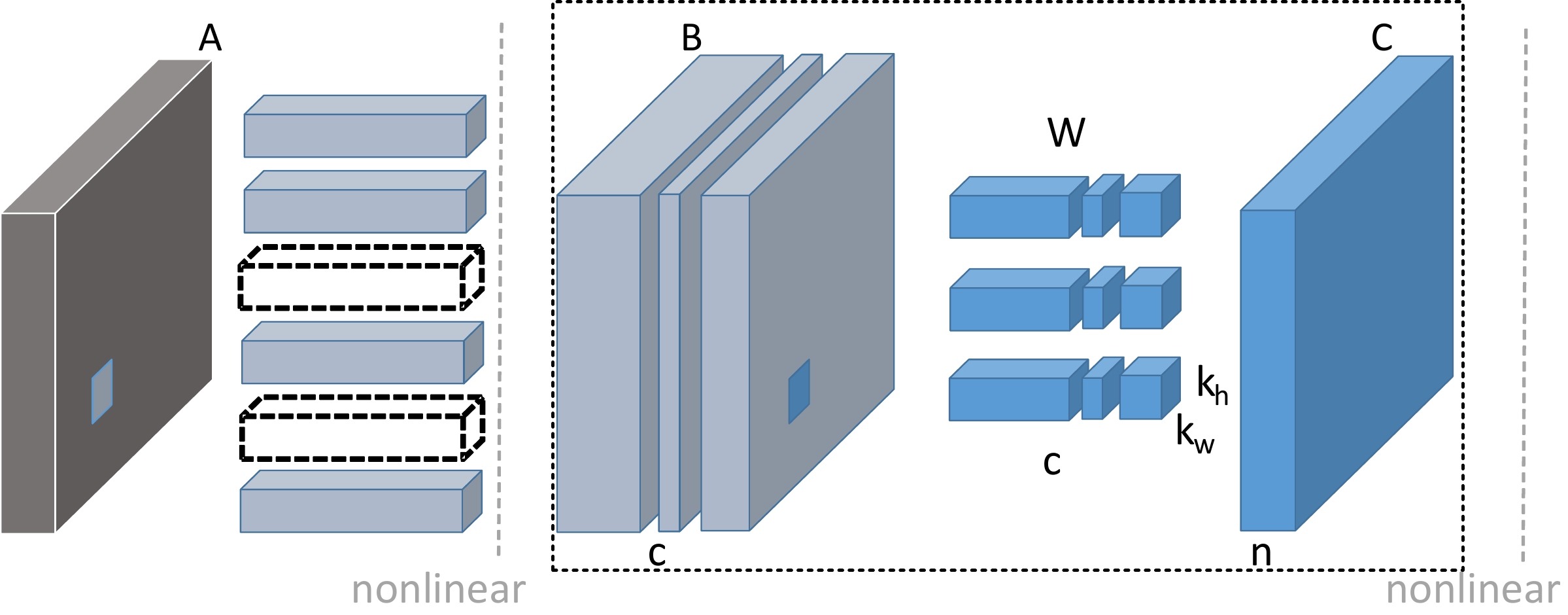}        
        \end{center}
        \caption{\Channelpruning\ for accelerating a \conv. 
        We aim to reduce the number of channels of feature map and minimize the reconstruction error on feature map C.
        Our optimization algorithm (Sec.~\ref{sec:linear}) performs within the dotted box, which does not involve nonlinearity.
        This figure illustrates the situation that two channels are pruned for feature map B. 
        Thus corresponding channels of filters $\wtwo$ can be removed.
        Furthermore, even though not directly optimized by our algorithm, 
        the corresponding filters in the previous layer can also be removed
        (marked by dotted filters).
         $c,n$: number of channels for feature maps B and C, $k_h\times k_w$: kernel size. 
        }
        \label{fig:ill}
    \end{figure*}

    In this paper, we propose a new inference-time approach for \channelpruning, 
    utilizing inter-channel redundancy.
    Inspired by \tensordecom\ improvement by feature maps reconstruction~\cite{Zhang2015}, 
    instead of pruning according to filter weights magnitude~\cite{Li2016,luo2017thinet}, 
    we fully exploit redundancy inter feature maps. 
    Instead of recovering performance with finetuning~\cite{han2015learning,Li2016,luo2017thinet}, we propose to reconstruct output after pruning each layer.
    Specifically, given a trained CNN model, pruning each layer is achieved by minimizing reconstruction error on its output feature maps, 
    as shown in Fig.~\ref{fig:ill}.
    We solve this minimization problem by two alternative steps: channels selection and feature map reconstruction.
    In one step, we figure out the most representative channels,
    and prune redundant ones, based on \las.
    In the other step, we reconstruct the outputs with remaining channels with linear least squares.
    We alternatively take two steps. Further, we approximate the network layer-by-layer, with accumulated error accounted.
    We also discuss methodologies to prune multi-branch networks (\eg, ResNet~\cite{He2015}, Xception~\cite{chollet2016xception}).
        
    We demonstrate the superior accuracy of our approach over other recent pruning techniques~\cite{lin2017runtime,luo2017thinet,wang2017structured,liu2017learning}. For VGG-16, we achieve \x{4} acceleration, with only \textbf{\vggfour\%} increase of top-5 error.
    Combined with \tensordecom, we reach \x{5} acceleration but merely suffer \textbf{\vggcft\%} increase of error, 
    which outperforms previous state-of-the-arts.
    We further speed up ResNet-50 and Xception-50 by \x{2} with only \textbf{\resft\%, \xceptionft\%} accuracy loss respectively. Code has been made publicly available \footnote{https://github.com/yihui-he/channel-pruning}.
    
    A preliminary version of this manuscript has been accepted to a conference~\cite{He_2017_ICCV}. This manuscript extends the initial version from several aspects to strengthen our approach:
    \begin{enumerate}
    \item We investigated inter-channel redundancy in each layer, and better analysis it for pruning. 
    \item We present filter-wise pruning, which has compelling acceleration performance for a single layer. 
    \item We demonstrated compelling VGG-16 acceleration Top-1 results which outperform its original model.
    \end{enumerate}

\section{Related Work}

    There has been a significant amount of work on accelerating CNNs~\cite{cheng2017survey}, 
    starting from brain damage~\cite{lecun1989optimal,hassibi1993second}.
    Many of them fall into three categories: \implementation~\cite{bagherinezhad2016lcnn}, 
    quantization~\cite{Rastegari2016}, and \structured~\cite{jaderberg2014speeding}.

    \Implementation\ based methods~\cite{mathieu2013fast,vasilache2014fast,lavin2015fast,bagherinezhad2016lcnn} accelerate convolution, with special convolution algorithms like FFT~\cite{vasilache2014fast}.
    Quantization~\cite{courbariaux2016binarynet,Rastegari2016,phan2020mobinet} reduces floating point computational complexity. 

    \Sparseconnect\ eliminates connections between neurons~\cite{han2015learning,liu2015sparse,lebedev2015fast,han2016eie,guo2016dynamic,zhong2018shift,he2019addressnet}.
     \cite{yang2016designing} prunes connections based on weights magnitude. 
     \cite{han2015deep} could accelerate fully connected layers up to \x{50}.
    However, in practice, the actual speed-up may be very related to implementation.

    \Tensordecom~\cite{jaderberg2014speeding,lebedev2014speeding,gong2014compressing,kim2015compression,he2019depth} decompose weights into several pieces.
    \cite{xue2013restructuring,denton2014exploiting,girshick2015fast} accelerate fully connected layers with truncated SVD.
    \cite{Zhang2015} factorize a layer into $3\times3$ and $1\times1$ combination, driven by feature map redundancy.

    \Channelpruning\ removes redundant channels on feature maps.
    There are several training-based approaches \cite{zhang2020image}. 
    \cite{Alvarez2016,wen2016learning,zhou2016less} regularize networks to improve accuracy.
    Channel-wise SSL~\cite{wen2016learning} reaches high compression ratio for first few conv layers of LeNet~\cite{lecun1998gradient} and AlexNet~\cite{krizhevsky2012imagenet}.
    \cite{zhou2016less} could work well for fully connected layers.
    However,
    \textit{training-based} approaches are more costly,
    and their effectiveness on very deep networks on large datasets is rarely exploited.
    Inference-time \channelpruning\ is challenging, as reported by previous works~\cite{anwar2015structured,polyak2015channel}. Recently, AMC~\cite{He_2018_ECCV} improves our approach by learning speed-up ratio with reinforcement learning. 

    Some works~\cite{srinivas2015data,mariet2015diversity,hu2016network} focus on model size compression, which mainly operate the fully connected layers.
    Data-free approaches~\cite{Li2016,anwar2016compact} results for \ratio\ (\eg, \x{5}) have not been reported,
    and requires long retraining procedure.
     \cite{anwar2016compact} select channels via over 100 random trials. However, it needs a long time to evaluate each trial on a deep network,
    which makes it infeasible to work on very deep models and large datasets.
    \cite{Li2016} is even worse than naive solution from our observation sometimes (Sec. \ref{sec:ablation}).

\section{Approach}
    
    In this section, we first propose a \channelpruning\ algorithm for a single layer,
    then generalize this approach to multiple layers or the whole model.
    Furthermore, we discuss variants of our approach for multi-branch networks.
    
    \subsection{Formulation}\label{sec:linear}
    \newcommand{\rankcontrain}{ \norm{\lcoef}_0 \leq \rankp}
    \newcommand{\wtwocontrain}{ \forall i \norm{\wtwoi}_F =1}    
    
    \figurename~\ref{fig:ill} illustrates our \channelpruning\ algorithm for a single \conv.
    We aim to reduce the number of channels of feature map B while maintaining outputs in feature map C.
    Once channels are pruned, 
    we can remove corresponding channels of the filters that take these channels as input.
    Also, filters that produce these channels can also be removed.
    It is clear that \channelpruning\ involves two key points. 
    The first is channel selection since we need to select proper channel combination to maintain as much information.
    The second is reconstruction. We need to reconstruct the following feature maps using the selected channels.
    
    Motivated by this, we propose an iterative two-step algorithm:
    \begin{enumerate}
    \item In one step, we aim to select most representative channels.
    Since an exhaustive search is infeasible even for small networks,
    we come up with a \las\ based method to figure out representative channels and prune redundant ones.
    \item In the other step, we reconstruct the outputs with remaining channels with linear least squares.
    \end{enumerate}
    We alternatively take two steps.

    Formally, to prune a feature map B with $\rank$ channels, we consider applying $\out \times \rank \times \kh \times \kw$ convolutional filters $\wtwo$ on $\samp \times \rank  \times \kh \times \kw$ input volumes $\reX$ sampled from this feature map, which produces $\samp \times \out$ output matrix $\reY$ from feature map C.
    Here, $\samp$ is the number of samples, $\out$ is the number of output channels, and $\kh, \kw$ are the kernel size. 
    For simple representation, bias term is not included in our formulation.
    To prune the input channels from $\rank$ to desired $\rankp$ ($0 \leq \rankp \leq \rank$), while minimizing reconstruction error, 
    we formulate our problem as follow:
    \begin{equation}\label{eq:l0}
    \begin{aligned}
    &     \argmin_{\lcoef, \wtwo} \frac{1}{2\samp} \norm{\reY - \sum_{i=1}^{\rank} \lcoefi \reXi \wtwoi^\top }^2_F \\
    & \st \rankcontrain
    \end{aligned}
    \end{equation}
    
    $\norm{ \cdot }_F$ is Frobenius norm.
    $\reXi$ is $\samp \times \kh\kw$ matrix sliced from $i$th channel of input volumes $\reX$, $i = 1,...,\rank$.
    $\wtwoi$ is $\out \times \kh\kw$ filter weights sliced from $i$th channel of $\wtwo$. 
    $\lcoef$ is coefficient vector of length $\rank$ for channel selection, and $\lcoefi$ ($i$th entry of $\lcoef$) is a scalar mask to $i$th channel (i.e. to drop the whole channel or not).
    Notice that, if $\lcoefi = 0$, $\reXi$ will be no longer useful,
    which could be safely pruned from feature map B.
    $\wtwoi$ could also be removed.
    $\rankp$ is the number of retained channels, which is manually set as it can be calculated from the desired speed-up ratio. For whole-model speed-up (i.e. Section \ref{sec:ratio}), given the overall speed-up, we first assign speed-up ratio for each layer then calculate each $\rankp$.
    
    \subsection{Optimization}
    Solving this $\ell_0$ minimization problem in Eqn.~\ref{eq:l0} is NP-hard. 
    Therefore, we relax the $\ell_0$ to $\ell_1$ regularization:
    \begin{equation}\label{eq:l1}
    \begin{aligned}
    &     \argmin_{\lcoef, \wtwo} \frac{1}{2\samp} \norm{\reY - \sum_{i=1}^{\rank} \lcoefi \reXi \wtwoi^\top }^2_F + \lalpha \norm{\lcoef}_1 \\
    & \st \rankcontrain, \wtwocontrain
    \end{aligned}
    \end{equation}
    
    $\lalpha$ is a penalty coefficient.
    By increasing $\lalpha$, there will be more zero terms in $\lcoef$ and one can get higher \ratio.
    We also add a constraint $\wtwocontrain$ to this formulation to avoid trivial solution.

    Now we solve this problem in two folds.
    First, we fix $\wtwo$, solve $\lcoef$ for channel selection.    
    Second, we fix $\lcoef$, solve $\wtwo$ to reconstruct error.
    
    \subsubsection{(i) The subproblem of $\lcoef$} \label{subprob0}
    In this case, $\wtwo$ is fixed. 
    We solve $\lcoef$ for channel selection.
    This problem can be solved by \las~\cite{tibshirani1996regression,breiman1995better}, which is widely used for model selection.
    \begin{equation}
    \begin{aligned}
    & \hat{\lcoef}^{LASSO}(\lalpha)= \argmin_{\lcoef} \frac{1}{2\samp}\norm{\reY -  \sum_{i=1}^{\rank} \lcoefi\wxi}_F^2 + \lalpha \norm{\lcoef}_1 \\
    & \st \rankcontrain
    \end{aligned}
    \end{equation}
    Here $\wxi = \reXi\wtwoi^\top$ (size $\samp \times \out$).
    We will ignore $i$th channels if $\lcoefi = 0$.
    
    \subsubsection{(ii) The subproblem of $\wtwo$} \label{subprob1}
    In this case, $\lcoef$ is fixed.
    We utilize the selected channels to minimize reconstruction error.
    We can find optimized solution by least squares:
    \begin{equation}
    \argmin_{\reW} \norm{\reY - \cx(\reW)^\top}_F^2
    \end{equation}
    Here $\cx = [\lcoefx{1}\reXx{1}\; \lcoefx{2}\reXx{2}\; ...\; \lcoefx{i}\reXx{i}\; ...\; \lcoefx{c}\reXx{c}]$ (size $\samp \times  \rank\kh\kw$). 
    $\reW$ is $\out \times \rank\kh\kw$ reshaped $\wtwo$, 
    $\reW=[\wtwox{1}\; \wtwox{2}\; ...\; \wtwox{i}\; ...\; \wtwox{c}]$.
    After obtained result $\reW$, it is reshaped back to $\wtwo$.
    Then we assign $\lcoefi \leftarrow \lcoefi \norm{\wtwoi}_F, \wtwoi \leftarrow \wtwoi / \norm{\wtwoi}_F$. 
    Constrain $\wtwocontrain$ satisfies.

    We alternatively optimize (i) and (ii).
    In the beginning, $\wtwo$ is initialized from the trained model,  $\lalpha=0$, namely no penalty,
    and $\norm{\lcoef}_0=\rank$.
    We gradually increase $\lalpha$. 
    For each change of $\lalpha$, we iterate these two steps until $\norm{\lcoef}_0$ is stable. 
    After $\rankcontrain$ satisfies, 
    we obtain the final solution $\wtwo$ from $\{\lcoefx{i}\wtwox{i}\}$.
    In practice, we found that the two steps iteration is time consuming. 
    So we apply (i) multiple times until $\rankcontrain$ satisfies.
    Then apply (ii) just once, to obtain the final result.
    From our observation, this result is comparable with two steps iteration's result.
    Therefore, in the following experiments, we adopt this approach for efficiency.

    \subsubsection{Discussion}
    Some recent works~\cite{wen2016learning,Alvarez2016,han2015learning} (though training based) also introduce $\ell_1$-norm or LASSO.
    However, we must emphasize that we use different formulations.
    Many of them introduced sparsity regularization into training loss, 
    instead of explicitly solving LASSO.
    Other work~\cite{Alvarez2016} solved LASSO, while feature maps or data were not considered during optimization.

    Because of these differences, our approach could be applied at inference time.
    
    \subsection{Whole Model Pruning}\label{sec:whole}
    Inspired by \cite{Zhang2015}, we apply our approach layer by layer sequentially.
    For each layer, we obtain input volumes from the current input feature map, 
    and output volumes from the output feature map of the un-pruned model.
    This could be formalized as: 

    \begin{equation}\label{eq:whole}
    \begin{aligned}
    &     \argmin_{\lcoef, \wtwo} \frac{1}{2\samp} \norm{\reY^\prime - \sum_{i=1}^{\rank} \lcoefi \reXi \wtwoi^\top }^2_F \\
    & \st \rankcontrain
    \end{aligned}
    \end{equation}
    
    Different from Eqn.~\ref{eq:l0}, $\reY$ is replaced by $\reY^\prime$, 
    which is from feature map of the original model.
    Therefore, the accumulated error could be accounted during sequential pruning.
    
    \subsection{Pruning Multi-Branch Networks}\label{sec:variants}
\begin{figure}
    \begin{center}
       \includegraphics[width=\linewidth]{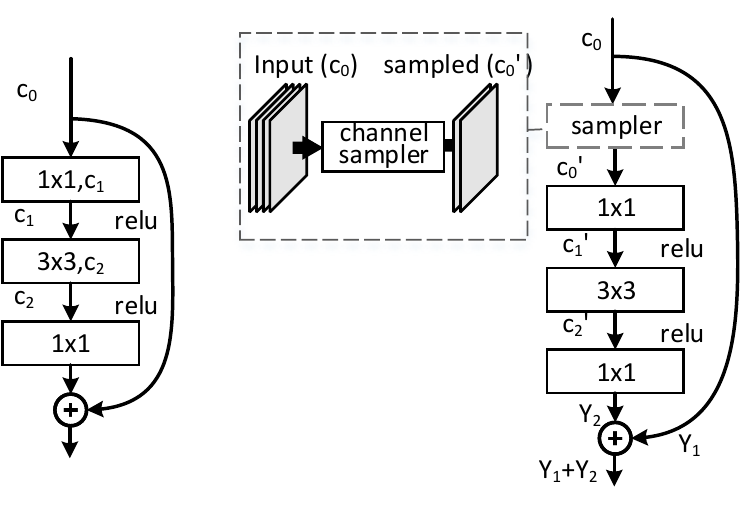}
    \end{center}
    \caption{Illustration of multi-branch enhancement for residual block. 
        \textbf{Left}: original residual block. 
        \textbf{Right}: pruned residual block with enhancement, $\mathrm{c_x}$ denotes the \featch. Input channels of the first \conv\ are sampled, so that the large input \featch\ could be reduced. As for the last layer, rather than approximate $\reY_2$, we try to approximate $\reY_1+\reY_2$ directly (Sec.~\ref{sec:multi} Last layer of residual branch).}
    \label{fig:sampler}
    \end{figure}

    The whole model pruning discussed above is enough for single-branch networks like LeNet~\cite{lecun1998gradient}, AlexNet~\cite{krizhevsky2012imagenet} and VGG Nets~\cite{Simonyan2014}.
    However, it is insufficient for multi-branch networks like GoogLeNet~\cite{szegedy2015going} and  ResNet~\cite{He2015}.
    We mainly focus on pruning the widely used residual structure (\eg, ResNet~\cite{He2015}, Xception~\cite{chollet2016xception}).
    Given a residual block shown in \figurename~\ref{fig:sampler}~(left), the input bifurcates into the shortcut and the residual branch.
    On the residual branch, there are several \conv s (\eg, 3 \conv s which have spatial size of $1\times1,3\times3,1\times1$, \figurename~\ref{fig:sampler}, left). 
    Other layers except the first and last layer can be pruned as is described previously.
    For the first layer, the challenge is that the large input \featch\ (for ResNet, four times of its output) can't be easily pruned since it is shared with the shortcut.
    For the last layer, accumulated error from the shortcut is hard to be recovered, since there's no parameter on the shortcut.
    To address these challenges, we propose several variants of our approach as follows.

    \label{sec:multi}
    \subsubsection{Last layer of residual branch}
    Shown in \figurename~\ref{fig:sampler}, the output layer of a residual block consists of two inputs: 
    feature map $\reY_1$ and $\reY_2$ from the shortcut and residual branch.
    We aim to recover $\reY_1 + \reY_2$ for this block.
    Here, $\reY_1, \reY_2$ are the original feature maps before pruning.
    $\reY_2$ could be approximated as in Eqn.~\ref{eq:l0}.
    However, shortcut branch is parameter-free, then $\reY_1$ could not be recovered directly.
    To compensate this error, 
    the optimization goal of the last layer is changed from $\reY_2$ to $\reY_1 - \reY^\prime_1 + \reY_2$,
    which does not change our optimization.
    Here, $\reY^\prime_1$ is the current feature map after previous layers pruned.
    When pruning, volumes should be sampled correspondingly from these two branches.

    \subsubsection{First layer of residual branch}
    Illustrated in \figurename~\ref{fig:sampler}(left),
    the input feature map of the residual block could not be pruned,
    since it is also shared with the shortcut branch.
    In this condition, we could perform \sampling\ before the first convolution to save computation.
    We still apply our algorithm as Eqn.~\ref{eq:l0}.
    Differently, we sample the selected channels on the shared feature maps to construct a new input for the later convolution,
    shown in \figurename~\ref{fig:sampler}(right).
    The computational cost for this operation could be ignored.    
    More importantly, after introducing \textit{feature map sampling},
    the convolution is still "regular".
    
    \begin{figure}
        \begin{center}
                \includegraphics[width=\linewidth]{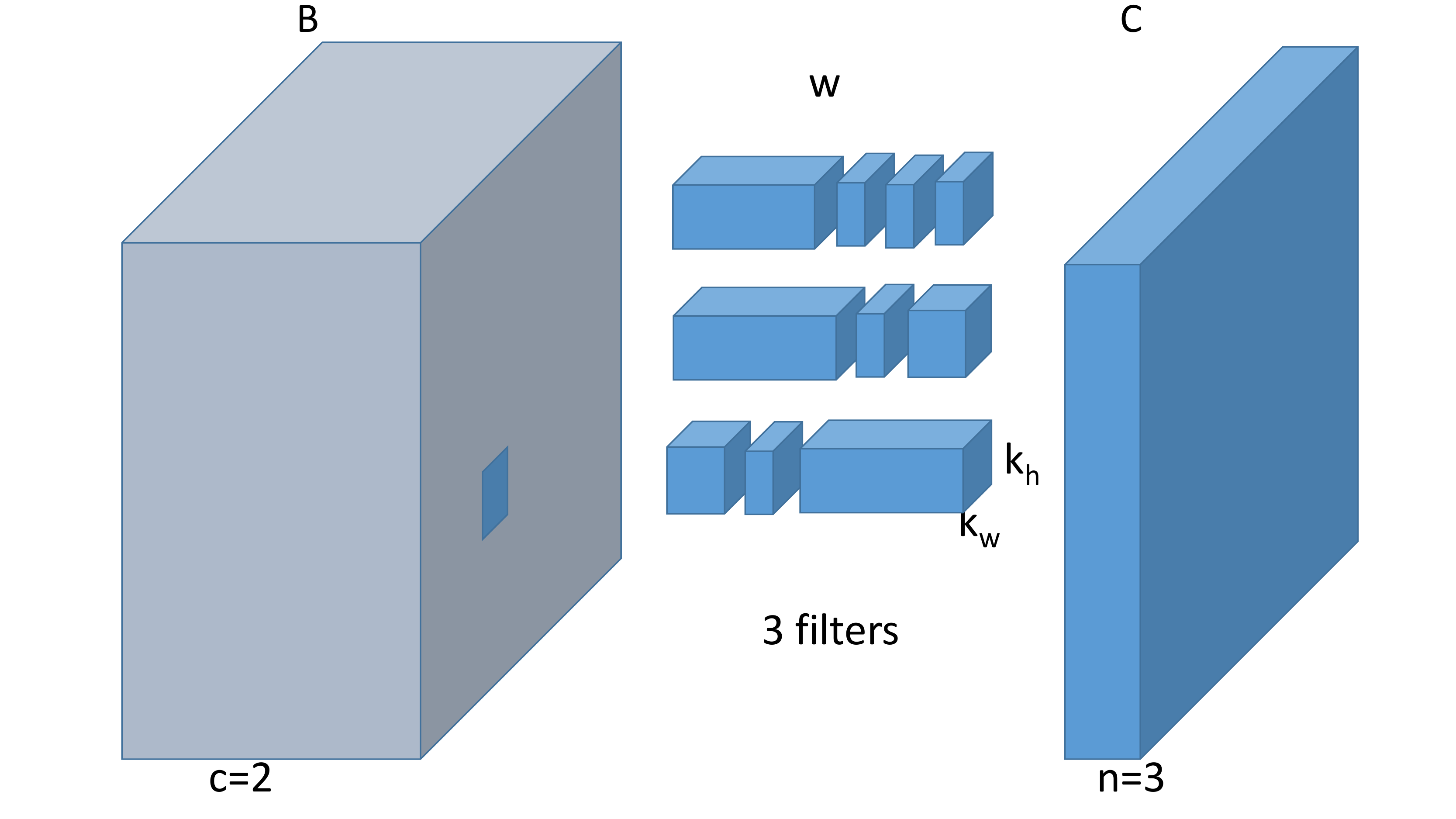}
        \end{center}
        \caption{\Filterwise\ for accelerating the first \conv\ on the residual branch. 
        We aim to reduce the number of channels filter-wise in weights $\wtwo$,
        while minimizing the reconstruction error on feature map C. Channels of feature map B is not pruned.
        We apply our Eqn.~\ref{eq:l0} to each filter independently (each filter chooses its own representative input channels).
         $c,n$: number of channels for feature maps B and C, $k_h\times k_w$: kernel size. 
        }
        \label{fig:filterwise}
    \end{figure}
    
    \Filterwise\ is another option for the first convolution on the residual branch, shown in \figurename~\ref{fig:filterwise}.
    Since the input channels of parameter-free shortcut branch could not be pruned,
    we apply our Eqn.~\ref{eq:l0} to each filter independently (each filter chooses its own representative input channels). It outputs "irregular" \conv s, which need special library implementation support.
    
        \subsection{Combined with \TensorDecom}\label{sec:3C}
    \Channelpruning\ can be easily combined method with \tensordecom, quantization, and lowbits etc.
    We focus on combination with \tensordecom.
    
        \begin{figure*}[!t]
    \begin{center}
        \includegraphics[width=\linewidth]{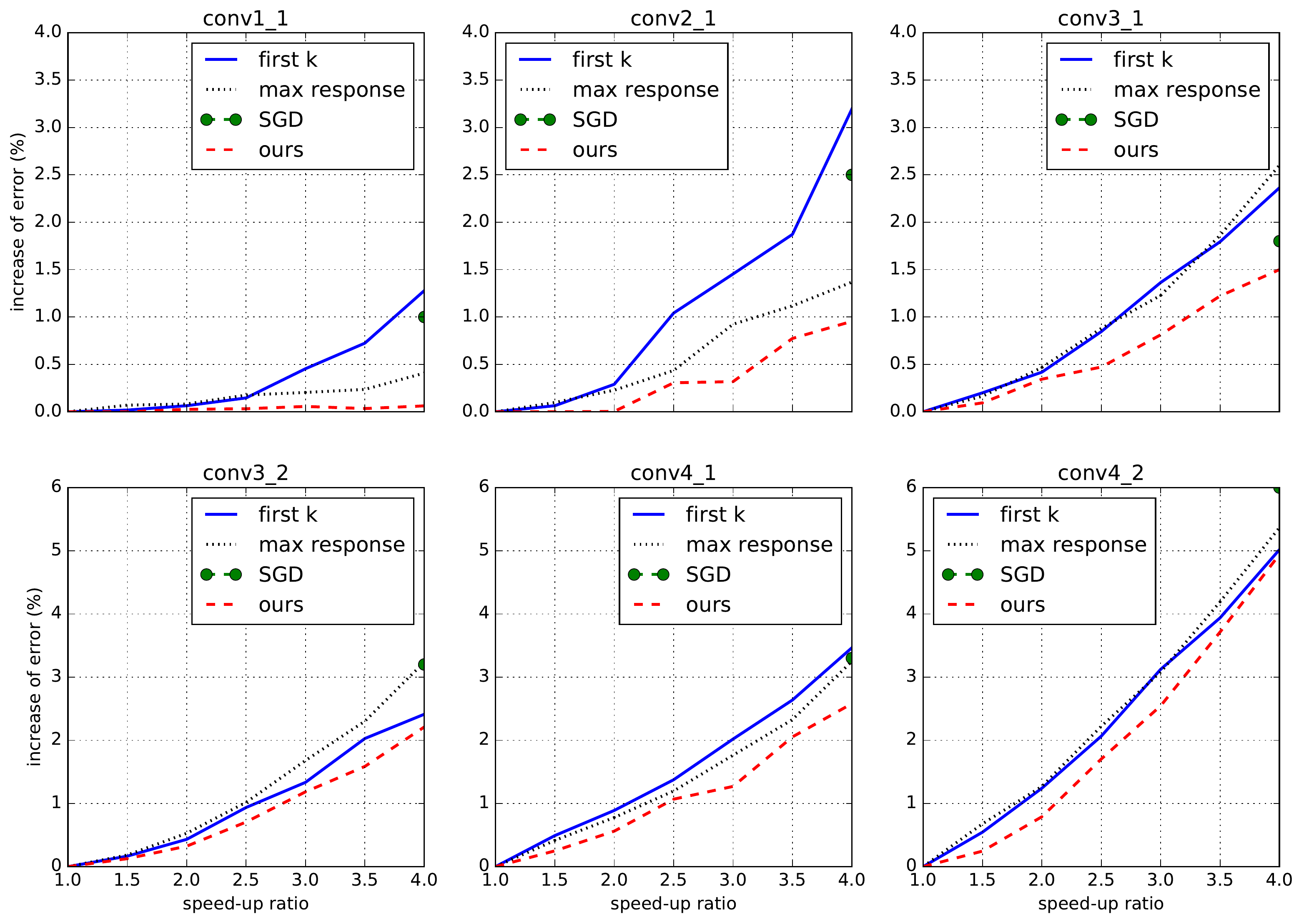}
    \end{center}
    \caption{Single layer performance analysis under different speed-up ratios (without fine-tuning), measured by increase of error.
        To verify the importance of channel selection refered in Sec.~\ref{sec:linear}, we considered three naive baselines.
         \firstk\ selects the first $k$ feature maps. \prune\ selects channels based on absolute sum of corresponding weights filter~\cite{Li2016}.
         \sgd\ is a simple SGD alternative of our approach. Our approach is consistently better (\textit{smaller is better}).}
    \label{fig:ablation}
    \end{figure*}

    In general, \tensordecom\ could be represent as:
    \begin{equation}
    W^{l_n} = W_1\cdot W_2\cdot...\cdot W_n
    \end{equation}
    Here, $W^{l_n}$ is the original \conv\ filters for layer $n$, and $W_1\cdot W_2\cdot...\cdot W_n$ are several decomposed weights of the same size as $W$.
    Since the input and output channels of \tensordecom\ methods could not shrink, 
    it becomes a bottleneck when reaching high \ratio. 
    We apply channels reduction on first and last weights decomposed layers, namely the output of $W_n$ and the input of $W_1$.
    In our experiments (Sec.~\ref{sec:3Cexp}), we combined \cite{jaderberg2014speeding}, \cite{Zhang2015} and our approach. First, a $3\times3$ weights is decomposed to $3\times1,1\times3,1\times1$. Then our approach is applied to $3\times1$ and $1\times1$ weights.

        \subsection{Fine-tuning}
    We fine-tune the approximated model end-to-end on training data,
    which could gain more accuracy after reduction. 
    We found that since the network is in a pretty unstable state, 
    fine-tuning is very sensitive to the learning rate. 
    The learning rate needs to be small enough. Otherwise, the accuracy quickly drops. 
    If the learning rate is large, the finetuning process may jump out of the initialized local optimum by the pruned network and behave very similar to training the pruned architecture from scratch (Table~\ref{tab:orig}).

    On ImageNet, we use learning rate of $1e^{-5}$ and a mini-batch size of 128. 
    Fine-tune the models for ten epochs in the Imagenet training data (1/12 iterations of training from scratch). 
    On CIFAR-10, we use learning rate of $1e^{-4}$ and a mini-batch size of 128 and fine-tune the models for 6000 iterations (training from scratch need 64000 iterations).

\section{Experiment}

    \begin{table*}
\caption{The VGG-16 architecture. The column "complexity" is portion of the theoretical time complexity each layer contributes. The column "PCA energy" shows feature map PCA Energy (top 50\% eigenvalues). }
\label{tab:pca}
\begin{center}
\begin{tabular}{|l|c|c|c|c|c|}
\hline
           & \# channels & \# filters & output size & complexity (\%) & PCA energy (\%) \\ \hline
conv1\_1  & 64          & 64         & 224         & 0.6             & 99.8            \\ 
conv1\_2  & 64          & 64         & 224         & 12              & 99.0            \\ 
pool1    &             &            & 112         &                 &                 \\ \hline
conv2\_1  & 64          & 128        & 112         & 6               & 96.7            \\ 
conv2\_2  & 128         & 128        & 112         & 12              & 92.9            \\ 
pool2    &             &            & 56          &                 &                 \\ \hline
conv3\_1  & 128         & 256        & 56          & 6               & 94.8            \\ 
conv3\_2  & 256         & 256        & 56          & 12              & 92.3            \\ 
conv3\_3  & 256         & 256        & 56          & 12              & 89.3            \\ 
pool3    &             &            & 28          &                 &                 \\ \hline
 conv4\_1 & 256         & 512        & 28          & 6               & 89.9            \\ 
conv4\_2  & 512         & 512        & 28          & 12              & 86.5            \\ 
conv4\_3  & 512         & 512        & 28          & 12              & 81.8            \\ 
pool4    &             &            & 14          &                 &                 \\ \hline
conv5\_1  & 512         & 512        & 14          & 3               & 83.4            \\ 
conv5\_2  & 512         & 512        & 14          & 3               & 83.1            \\ 
conv5\_3  & 512         & 512        & 14          & 3               & 80.8            \\ \hline
\end{tabular}
\end{center}
\end{table*}
    
\begin{table*}[]
\caption{Accelerating the VGG-16 model~\cite{Simonyan2014} using a speedup ratio of \x{2}, \x{4}, or \x{5} (\textit{smaller is better}).}
        \label{tab:theo}
\begin{center}        
\begin{tabular}{|c|c|c|c|}
\hline
\multicolumn{4}{|c|}{Increase of top-5 error (1-view, baseline \origvgg\%)} \\ \hline
Solution                    & \x{2}         & \x{4}         & \x{5}         \\ \hline
\jader                      & -             & 9.7           & 29.7          \\ \hline
\asym                       & 0.28          & 3.84          & -             \\ \hline
\prft                   & 0.8           & 8.6           & 14.6          \\ \hline
\RNP    &  -   & 3.23   &3.58   \\ \hline
\spp    &  0.3   & 1.1   &2.3   \\ \hline
Ours (without fine-tune)    & \vggtworaw    & \vggfourraw   & \vggfiveraw   \\ \hline
Ours (fine-tuned)           & \textbf{0}             & \textbf{\vggfour}      & \textbf{\vggfive}      \\ \hline

\end{tabular}
\end{center}
\end{table*}

    We evaluation our approach for the popular VGG Nets~\cite{Simonyan2014}, ResNet~\cite{He2015}, Xception~\cite{chollet2016xception} on ImageNet~\cite{JiaDeng2009}, CIFAR-10~\cite{Krizhevsky2009} and PASCAL VOC 2007~\cite{pascal-voc-2007}.
    
    For Batch Normalization~\cite{ioffe2015batch}, 
    we first merge it into convolutional weights, 
    which do not affect the outputs of the networks. So that each \conv\ is followed by ReLU~\cite{nair2010rectified}.
    We use Caffe~\cite{jia2014caffe}\footnote{https://github.com/yihui-he/caffe-pro/tree/master} for deep network evaluation, 
    TensorFlow~\cite{tensorflow} for \sgd\ implementation (Sec.~\ref{sec:ablation})
    and scikit-learn~\cite{scikit-learn} for solvers implementation.
    For \channelpruning, we found that it is enough to extract 5000 images, and ten samples per image, which is also efficient (i.e., several minutes for VGG-16 \footnote{On Intel Xeon E5-2670 CPU}, Sec.~\ref{sec:ratio}).
    On ImageNet, we evaluate the top-5 accuracy with the single view. 
    Images are resized such that the shorter side is 256.
    The testing is on the center crop of $224\times224$ pixels.
    The augmentation for fine-tuning is the random crop of $224\times224$ and mirror.
    
    \subsection{Experiments with VGG-16}
    VGG-16~\cite{Simonyan2014} is a 16 layers single-branch convolutional neural network, with 13 \conv s. 
    It is widely used for recognition, detection and segmentation, \etc.
    Single view top-5 accuracy for VGG-16 is \origvgg\%\footnote{http://www.vlfeat.org/matconvnet/pretrained/}. 
    
    \subsubsection{Single Layer Pruning}\label{sec:ablation}\label{sec:filtersingle}
    
    In this subsection,  we evaluate single layer acceleration performance using our algorithm in Sec.~\ref{sec:linear}.
    For better understanding, we compare our algorithm with there naive channel selection strategies.
    \firstk\ selects the first \textit{k} channels. 
    \prune\ selects channels based on corresponding filters that have high absolute weights sum~\cite{Li2016}.
    \sgd\ is a simple alternative of our approach to use the original weights as initialization, as solve the $\ell_1$ regularized problem in Eqn.~\ref{eq:l1} (\textit{w.r.t.} both the weights and connections) by stochastic gradient descent.

    For fair comparison, 
    we obtain the feature map indexes selected by each of them, 
    then perform reconstruction (except \sgd, Sec. \ref{subprob1}).
    We hope that this could demonstrate the importance of channel selection.
    Performance is measured by the increase of error after a certain layer is pruned without fine-tuning, shown in Fig.~\ref{fig:ablation}.
     
    As expected, error increases as \ratio\ increases.
    Our approach is consistently better than other approaches in different \conv s under different \ratio.
    Unexpectedly, sometimes \prune\ is even worse than \firstk. 
    We argue that \prune\ ignores correlations between different filters. 
    Filters with large absolute weight may have a strong correlation. 
    Thus selection based on filter weights is less meaningful.
    Correlation on feature maps is worth exploiting.
    We can find that channel selection affects reconstruction error a lot. Therefore, it is important for \channelpruning. 
    
    As for \sgd, we only performed experiments under \x{4} speed-up due to time limitation. Though it shares same optimization goal with our approach, simple \sgd\ seems difficult to optimize to an ideal local minimal. Shown in \figurename~\ref{fig:ablation}, SGD is obviously worse than our optimization method. 
    
    Also notice that \channelpruning\ gradually becomes hard, from shallower to deeper layers. 
    It indicates that shallower layers have much more redundancy,
    which is consistent with \cite{Zhang2015}.
    We could prune more aggressively on shallower layers in whole model acceleration.
    
    \subsubsection{Whole Model Pruning} \label{sec:ratio}



    \begin{figure*}
    \begin{center}
        \includegraphics[width=\linewidth]{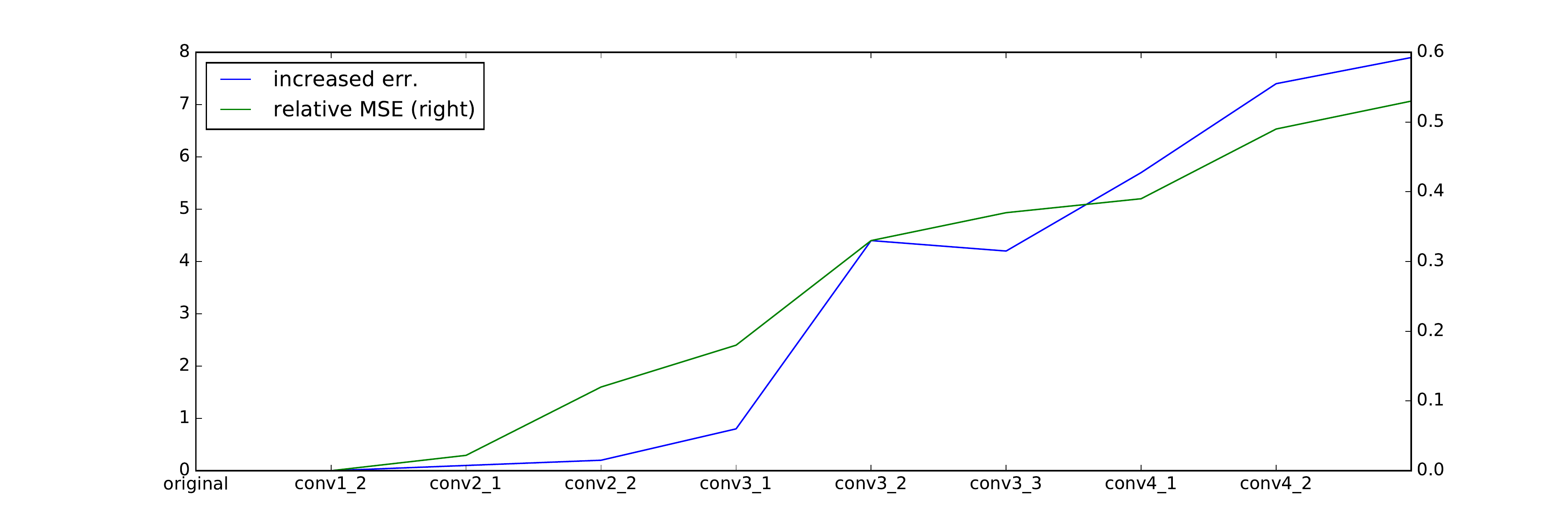}
    \end{center}
    \caption{Accumulated layerwise pruning error for accelerating VGG-16 under $4\times$. "relative MSE" is the relative mean square error. After fine-tuning, the Top-5 drops is 1.0\%.}
    \label{tab:accum}
    \end{figure*}
    

    \begin{figure*}
    \begin{center}
    \includegraphics[width=\linewidth]{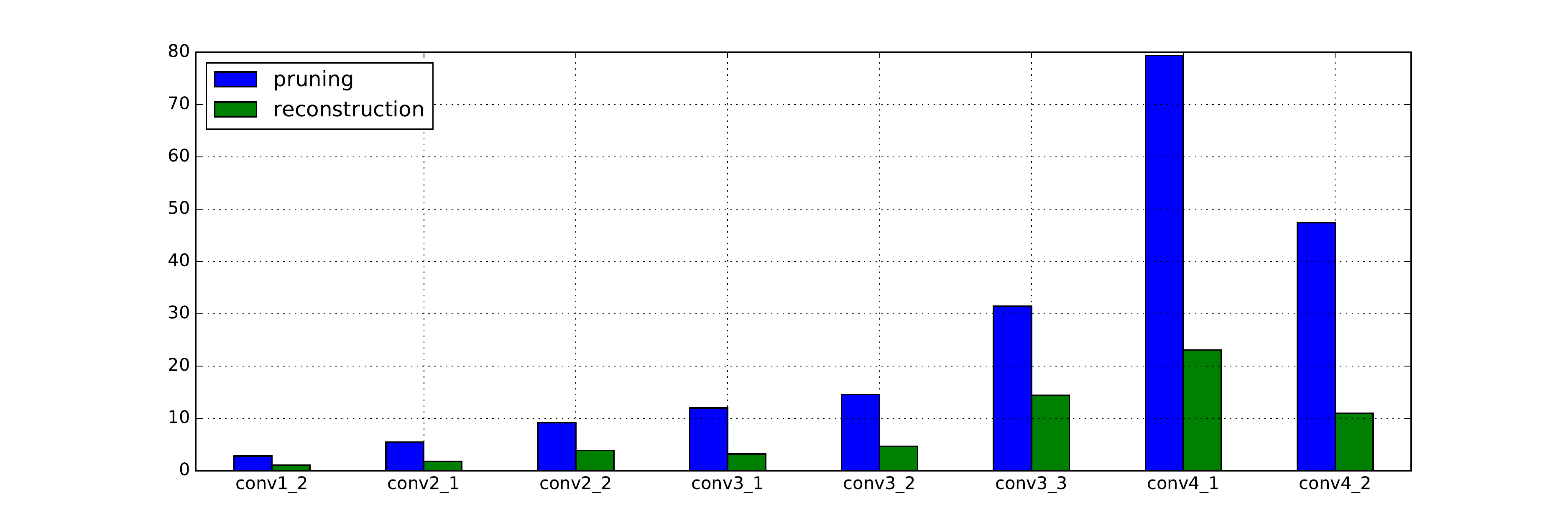}
    \end{center}
    \caption{Time consumption for pruning VGG-16 under \x{4}, on Intel Xeon E5-2670 CPU (measured by seconds, time consumption may vary a little in each run). Our algorithm is very efficient.}
    \label{tab:time}
    \end{figure*}

    Shown in Table~\ref{tab:pca}, we analyzed PCA energy of VGG-16. It indicates that shallower layers of VGG-16 are more redundant, which coincides with our single layer experiments above. So we prune more aggressive for shallower layers. Preserving channels ratios for shallow layers (\verb|conv1_x| to \verb|conv3_x|) and deep layers (\verb|conv4_x|) is $1:1.5$.
    \verb|conv5_x| are not pruned, since they only contribute 9\% computation in total and are not redundant, shown in Table~\ref{tab:pca}.
    
    We adopt the layer-by-layer whole model pruning proposed in Sec.~\ref{sec:whole}. \figurename~\ref{tab:accum} shows pruning VGG-16 under \x{4} speed-up, which finally reach 1.0\% increased of error after fine-tuning. It's easy to see that accumulated error grows layer-by-layer. And errors are mainly introduced by pruning latter layers, which coincides with our observation from single layer pruning and PCA analysis.
    
    Apart from the efficient inference model we attained, our algorithm is also efficient. Shown in \figurename~\ref{tab:time}, our algorithm could finish pruning VGG-16 under \x{4} within \textbf{5} minutes.
    
    Shown in Table~\ref{tab:theo}, whole model acceleration results under \x{2}, \x{4}, \x{5} are demonstrated.
    After fine-tuning, we could reach \x{2} speed-up without losing accuracy.
    Under \x{4}, we only suffers \vggfour\% drops.
    Consistent with single layer analysis, our approach outperforms other recent pruning approaches (Filter Pruning~\cite{Li2016}, Runtime Neural Pruning~\cite{lin2017runtime} and Structured Probabilistic Pruning~\cite{wang2017structured}) by a large margin.
    This is because we fully exploit channel redundancy within feature maps.
    Compared with \tensordecom\ algorithms, our approach is better than \jadercite,
    without fine-tuning. 
    Though worse than \asym,
    our combined model outperforms its combined Asym. 3D (Table~\ref{tab:combine}).
    This may indicate that \channelpruning\ is more challenging than \tensordecom, 
    since removing channels in one layer might dramatically change the input of the following layer.
    However, \channelpruning\ keeps the original model architecture, do not introduce additional layers, 
    and the absolute \ratio\ on GPU is much higher (Table~\ref{fig:real}).
    
    \subsubsection{Combined with Orthogonal Approaches}\label{sec:3Cexp}

    \begin{table}
        \caption{Performance of combined methods on the VGG-16 model~\cite{Simonyan2014} using a \ratio\ of \x{4} or \x{5}. 
        Our 3C solution outperforms previous approaches. The top-5 error rate
(1-view) of the baseline VGG-16 model is 10.1\%. (\textit{smaller is better}).}
        \begin{center}
    \begin{tabular}{|c|c|c|}
        \hline 
        \multicolumn{3}{|c|}{Increase of top-5 error (1-view)}                              \\ \hline
        Solution & \x{4} & \x{5}  \\ 
        \hline 
        \asymd & 0.9 & 2.0  \\ 
        \hline
        \asymdft & 0.3 & 1.0  \\ 
        \hline
        Our 3C & \vggcfour  & \vggc  \\ 
        \hline 
        Our 3C (fine-tuned) & \textbf{\vggcftfour} & \textbf{\vggcft}  \\ 
        \hline            
    \end{tabular} 
    \end{center}
    \label{tab:combine}
    \end{table}
    
    Since our approach exploits a new cardinality,
    we further combine our \channelpruning\ with spatial factorization~\cite{jaderberg2014speeding} and channel factorization~\cite{Zhang2015} (Sec~\ref{sec:3C}).
    Demonstrated in Table~\ref{tab:combine}, our 3 cardinalities acceleration (spatial, channel factorization, and \channelpruning, denoted by 3C) outperforms previous state-of-the-arts.
    \asymd\ (spatial and channel factorization), factorizes a \conv\ to three parts: $1\times3,\ 3\times1,\ 1\times1$.
    
    We apply spatial factorization, channel factorization, and our \channelpruning\ together sequentially layer-by-layer. 
    We fine-tune the accelerated models for 20 epochs, 
    since they are three times deeper than the original ones.
    After fine-tuning, our \x{4} model suffers no degradation.
    Clearly, a combination of different acceleration techniques is better than any single one.
    This indicates that a model is redundant in each cardinality.
    
\subsubsection{Performance without Output Reconstruction}
We evaluate whole model pruning performance without output reconstruction, to verify the effectiveness of the subproblem of $\wtwo$ (Sec.~\ref{subprob1}).
Shown in Table \ref{tab:lls}, without reconstruction, the accumulated error will be unacceptable for multi-layer pruning. Without reconstruction, the error increases to 99\%. Even after fine-tuning the score is still much worse than the counterparts. This is because the LASSO step (Sec.~\ref{subprob0}) only updates $\lcoef$ with limited freedom ($dimensionality = \rank $), thus it is insufficient for reconstruction. So we must adapt original weights $\wtwo$ ($\out \times \rank \times \kh \times \kw$) to the pruned input channels.

\begin{table}
    \caption{Accelerating VGG-16 under \x{4} with or without subproblem $\lcoef$. It's clear that without reconstruction the error increases. The top-5 error rate
(1-view) of the baseline VGG-16 model is 10.1\%.}
    \begin{center}
    \begin{tabular}{|c|c|c|}
        \hline
        \multicolumn{3}{|c|}{Increase of top-5 err. (1-view)} \\ \hline
        approach & before fine-tuning & fine-tuned \\ \hline
        Ours & 7.9 & 1.0 \\ \hline
        Without subproblem $\lcoef$ & 99 & 3.6 \\ \hline
    \end{tabular}
    \end{center}
    \label{tab:lls}
\end{table}

\subsubsection{Comparisons with Training from Scratch}\label{sec:vggscratch}
    \begin{table}
    \caption{Comparisons with training from scratch, under \x{4} acceleration on the VGG-16 model. Our fine-tuned model outperforms scratch trained counterparts. The top-5 error rate
(1-view) of the baseline VGG-16 model is 10.1\%. (\textit{smaller is better}).}
    \begin{center}
    \begin{tabular}{|c|c|c|}
        \hline 
           & Top-5 err. & Increased err.  \\ 
        \hline 
        From scratch  & \vggscratchacc  &  \vggscratcherr    \\ 
        \hline 
        From scratch (uniformed) & \vggscratchuniacc  & \vggscratchunierr     \\ 
        \hline 
        Ours & \vggfourrawacc & \vggfourraw  \\ 
        \hline
        Ours (fine-tuned) & \vggfouracc & \textbf{\vggfour}  \\ 
        \hline
    \end{tabular} 
    \end{center}
    \label{tab:orig}
    \end{table}

    Though training a compact model from scratch is time-consuming (usually 120 epochs),
    it worths comparing our approach and from scratch counterparts.
    To be fair, we evaluated both from scratch counterpart, 
    and normal setting network that has the same computational complexity and same architecture.

    Shown in Table~\ref{tab:orig}, 
    we observed that it's difficult for from scratch counterparts to reach competitive accuracy.
    Our model outperforms from scratch one.
    Our approach successfully picks out informative channels and constructs highly compact models.  
    We can safely draw the conclusion that the same model is difficult to be obtained from scratch.
    This coincides with architecture design studies~\cite{huang2016speed,Alvarez2016} 
    that the model could be easier to train if there are more channels in shallower layers.
    However, \channelpruning\ favors shallower layers.
    
    For from scratch (uniformed),
    the filters in each layer are reduced by half (e.g., reduce \verb|conv1_1| from 64 to 32).
    We can observe that normal setting networks of the same complexity couldn't reach the same accuracy either.
    This consolidates our idea that there's much redundancy in networks while training.
    However, redundancy can be opt-out at inference time.
    This may be an advantage of inference-time acceleration approaches over training-based approaches.
    
    Notice that there's a 0.6\% gap between the from scratch model and the uniformed one,
    which indicates that there's room for model exploration.
    Adopting our approach is much faster than training a model from scratch, even for a thinner one.
    Further researches could alleviate our approach to do thin model exploring.

  \subsubsection{Top-1 vs Top-5 Accuracy}
    \begin{table*}
    \caption{Increase of Top-1 and Top-5 error for accelerating VGG-16 on ImageNet. VGG-16 model's Top-5 and Top-1 error baselines are 29.5\% and 10.1\% respectively.}
    \begin{center}
    \begin{tabular}{|c|l|c|l|c|}
\hline
Model                     & \multicolumn{2}{l|}{Top-1}                 & \multicolumn{2}{l|}{Top-5}                 \\ \hline
\multicolumn{1}{|l|}{}    & err. & \multicolumn{1}{l|}{increased err.} & err. & \multicolumn{1}{l|}{increased err.} \\ \hline
$2\times$, fine-tuned     & 29.5 & 0.0                                 & 10.1 & 0.0                                 \\ \hline
$4\times$, fine-tuned     & 31.7 & 2.2                                 & 11.1 & 1.0                                 \\ \hline
$5\times$, fine-tuned     & 32.4 & 2.9                                 & 11.8 & 1.7                                 \\ \hline
3C, $4\times$, fine-tuned & 29.2 & -0.3                                & 10.1 & 0.0                                 \\ \hline
3C, $5\times$, fine-tuned & 29.5 & 0.0                                 & 10.4 & 0.3                                 \\ \hline
From scratch              & 31.9 & 2.4                                 & 11.9 & 1.8                                 \\ \hline
From scratch (uniformed)  & 32.9 & 3.4                                 & 12.5 & 2.4                                 \\ \hline
\end{tabular}
\end{center}
    \label{tab:top1}
\end{table*}

Though our approach already achieved good performance with Top-5 accuracy, it is still possible that it can lead to significant Top-1 accuracy decrease.
Shown in Table \ref{tab:top1}, we compare increase of Top-1 and Top-5 error for accelerating VGG-16 on ImageNet. 
Though the absolute drop is slightly larger, Top-1 is still consistent with top-5 results. For 3C \x{4} and \x{5}, the Top-1 accuracy is even better. 3C \x{4} Top-1 accuracy outperforms the original VGG-16 model by \textbf{0.3}\%. 

\subsubsection{Comparisons of Absolute Performance}\label{sec:gpu}
\begin{table*}[]
    \caption{GPU acceleration comparison. We measure forward-pass time per image. Our approach generalizes well on GPU. The top-5 error rate
(1-view) of the baseline VGG-16 model is 10.1\%. (\textit{smaller is better}).}
    \begin{center}
\begin{tabular}{|c|c|c|c|}
\hline
Model & Solution & Increased err. & GPU time/ms \\ \hline
VGG-16 & - & 0 & \round{3}{8.144125} \\ \hline
\multirow{6}{*}{VGG-16 ($4\times$)} & \jader & 9.7 & \round{3}{8.05059375} (\x{1.01}) \\ \cline{2-4} 
 & \asym & 3.8 & \round{3}{5.243625} (\x{1.55}) \\ \cline{2-4} 
 & \asymd & 0.9 & \round{3}{8.5031875} (\x{0.96}) \\ \cline{2-4} 
 & \asymdft & 0.3 & \round{3}{8.5031875} (\x{0.96}) \\ \cline{2-4} 
 & Ours (fine-tuned) & \vggfour & \textbf{3.264 (\x{2.50})} \\ \cline{2-4} 
 & Our 3C (fine-tuned) & \textbf{0.0} & 3.712 (\x{2.19}) \\ \hline
\end{tabular}
    \end{center}
    \label{fig:real}
\end{table*}

    We further evaluate absolute performance of acceleration on GPU.
    Results in Table~\ref{fig:real} are obtained under Caffe~\cite{jia2014caffe}, 
    CUDA8~\cite{cuda} and cuDNN5~\cite{cudnn}, with a mini-batch of 32 on a GPU\footnote{GeForce GTX TITAN X GPU}.
    Results are averaged from 50 runs.
    \Tensordecom\ approaches decompose weights into too many pieces, which heavily increase overhead.
    They could not gain much absolute speed-up.
    Though our approach also encountered performance decadence, 
    it generalizes better on GPU than other approaches.
    Our results for \tensordecom\ differ from previous research~\cite{Zhang2015,jaderberg2014speeding}, 
    maybe because current library and hardware prefer single large convolution instead of several small ones.

    \subsubsection{Acceleration for Detection}    
    VGG-16 is popular among object detection tasks~\cite{yolo,ssd,zhu2019feature,he2020deep,he2019bounding,he2019deep,he2018softer}.
    We evaluate transfer learning ability of our \x{2}/\x{4} pruned VGG-16,
    for Faster R-CNN~\cite{fasterrcnn}\footnote{https://github.com/rbgirshick/py-faster-rcnn} object detections.
    PASCAL VOC 2007 object detection benchmark~\cite{pascal-voc-2007} contains 5k trainval images and 5k test images.
    The performance is evaluated by mean Average Precision (mAP) and mmAP (AP at IoU=.50:.05:.95, primary challenge metric of COCO~\cite{lin2014microsoft}).
    In our experiments, we first perform \channelpruning\ for VGG-16 on the ImageNet. 
    Then we use the pruned model as the pre-trained model for Faster R-CNN.

    

\begin{table*}
\caption{\x{2}, \x{4} acceleration for Faster R-CNN detection. mmAP is AP at IoU=.50:.05:.95 (primary challenge metric of COCO~\cite{lin2014microsoft}). }
  \begin{center}
  \tabcolsep=0.11cm
\begin{tabular}{|c|c|c|c|c|c|c|c|c|c|c|c|c|c|c|c|c|c|c|c|c|c|c|}
\hline 
 & aero & bike & bird & boat & bottle & bus & car & cat & chair & cow & table & dog  & horse  & mbike & person & plant & sheep & sofa &train  & tv & mAP & mmAP \\ 
\hline 
orig & 68.5 & 78.1 & 67.8 & 56.6 & 54.3 & 75.6 & 80.0 & 80.0 & 49.4 & 73.7 & 65.5 & 77.7 & 80.6 & 69.9 & 77.5 & 40.9 & 67.6 & 64.6 & 74.7 & 71.7 & 68.7 & 36.7  \\ 
\hline 
\x{2} & 69.6&  79.3&  66.2&  56.1&  47.6&  76.8&  79.9&  78.2&  50.0&  73.0&  68.2&  75.7&  80.5&  74.8&  76.8&  39.1&  67.3&  66.8&  74.5&  66.1& 68.3 & 36.7\\ 
\hline 
\x{4} & 67.8 & 79.1 & 63.6 & 52.0 & 47.4 & 78.1 & 79.3 & 77.7 & 48.3 & 70.6 & 64.4 & 72.8 & 79.4 & 74.0 & 75.9 & 36.7 & 65.1 & 65.1 & 76.1 & 64.6 & 66.9 & 35.1 \\ 
\hline 
\end{tabular} 
\end{center}
\label{tab:det}
\end{table*}

    The actual running time of Faster R-CNN is 220ms / image. 
    The \conv s contributes about 64\%.
    We got actual time of 94ms for \x{4} acceleration.
   From Table~\ref{tab:det}, we observe 0.4\% mAP and 0.0\% mmAP drops of our \x{2} model, which is not harmful for practice consideration. Observed from mmAP, For higher localization requirements our speedup model does not suffer from large degradation.
    
    \subsection{Experiments with Residual Architecture Nets}
    For Multi-path networks~\cite{szegedy2015going,He2015,chollet2016xception}, 
    we further explore the popular ResNet~\cite{He2015} and latest Xception~\cite{chollet2016xception}, 
    on ImageNet and CIFAR-10.
    Pruning residual architecture nets is more challenging. 
    These networks are designed for both efficiency and high accuracy.
    \Tensordecom\ algorithms~\cite{Zhang2015,jaderberg2014speeding} are not applicable to these model.
    Spatially, $1\times1$ convolution is favored, which could hardly be factorized.

    \subsubsection{Filter-wise Pruning}
    
    \begin{figure}
    \begin{center}
    \includegraphics[width=\linewidth]{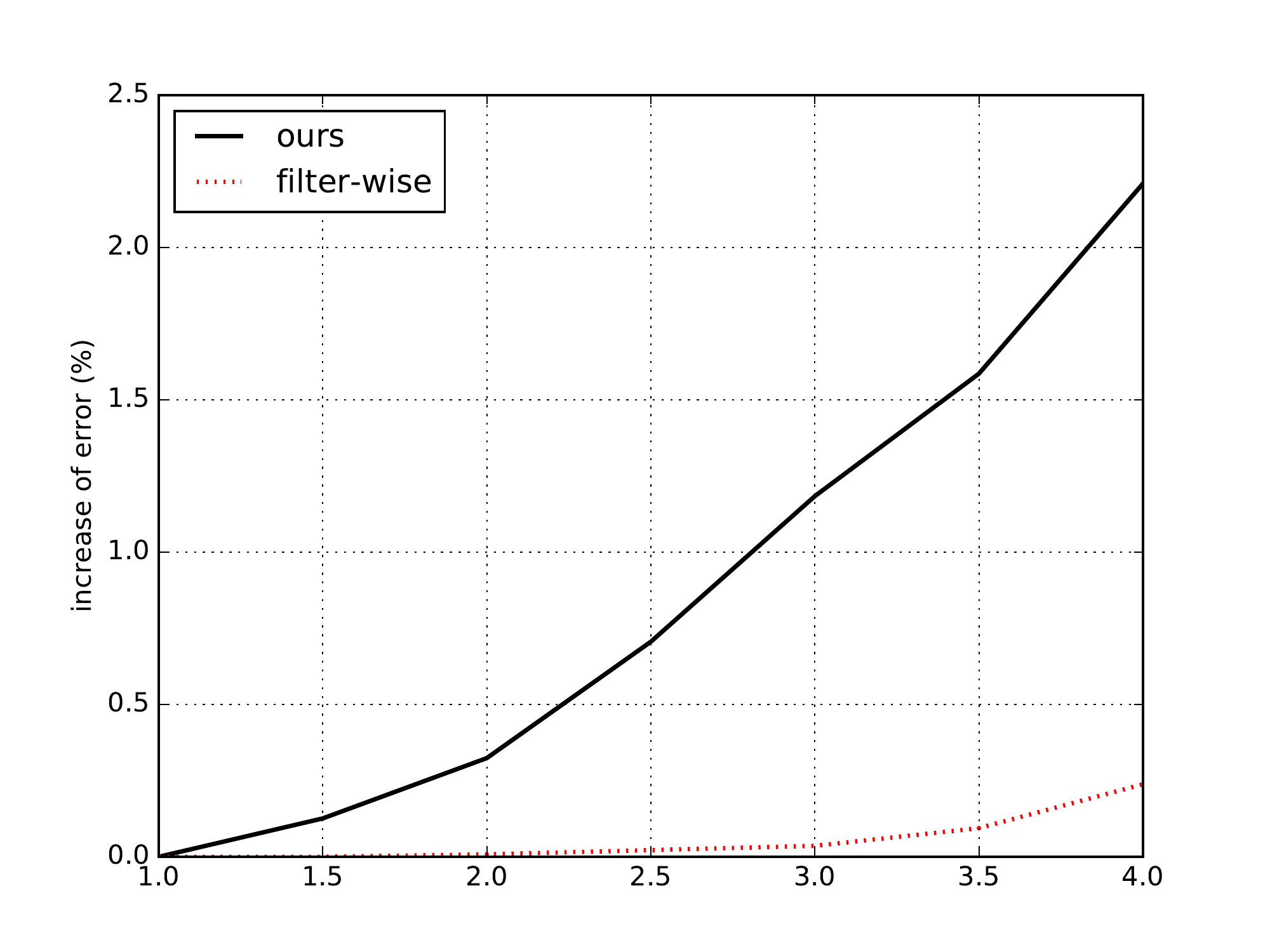}
    \end{center}
    \caption{\filterwise\ performance analysis under different speed-up ratios (without fine-tuning), measured by increase of error on VGG-16 conv3\_2.
        (\textit{smaller is better}).}
    \label{fig:filterperf}
    \end{figure}
    
    Under single layer acceleration, \filterwise\ (Sec.~\ref{sec:multi}) is more accurate than our original one, since it is more flexible, shown in \label{fig:filterperf}.
    From our ResNet pruning experiments in the next section, 
    it improves 0.5\% top-5 accuracy for \x{2} ResNet-50 (applied on the first layer of each residual branch) without fine-tuning.
    However, after fine-tuning, there's no noticeable improvement.
    Besides, it outputs "irregular" \conv s, which need special library implementation support to gain practical speed-up.
    Thus, we did not adopt it for our residual architecture nets.
    
    \subsubsection{ResNet Pruning} \label{sec:resnetimagenet}
    
\begin{table}[]
    \caption{ResNet-50 Computational Complexity (bottleneck structure). ResNet-50 complexity uniformly drops on each residual block. $ix$ stands for the $x$th block of $i$th stage, $i=2,3,4,5, x=a,b,c$. The "Complexity" column is the portion of computation complexity each block contributes.}
\begin{center}
\begin{tabular}{|c|c|}
\hline
layer name & Complexity (\textperthousand) \\ \hline
conv1      & 30                            \\ \hline
$2a_{1}$   & 13                            \\ \hline
$2a_{2a}$  & 3                             \\ \hline
$ia_{1}$   & 26                            \\ \hline
$ia_{2a}$  & 6                             \\ \hline
$ix_{2a}$  & 13                            \\ \hline
$ix_{2b}$  & 29                            \\ \hline
$ix_{2c}$  & 13                            \\ \hline
\end{tabular}
\end{center}
    \label{tab:rescomp}
\end{table}

    \begin{table}[]
        \caption{\x{2} acceleration for ResNet-50 on ImageNet, the baseline network's top-5 accuracy is \resorig\% (one view). 
        We improve performance with multi-branch enhancement (Sec. \ref{sec:multi}, \textit{smaller is better}).}
        \begin{center}
\begin{tabular}{|c|c|c|}
\hline
Solution & Speedup & Increased err. \\ \hline
\thinet & $1.58\times$ & 1.53 \\ \hline
\spp & $2\times$ & 1.8 \\ \hline
Ours & \multirow{3}{*}{2$\times$} & 8.0 \\ \cline{1-1} \cline{3-3} 
\begin{tabular}[c]{@{}c@{}}Ours\\ (enhanced)\end{tabular} &  & \resmb \\ \cline{1-1} \cline{3-3} 
\begin{tabular}[c]{@{}c@{}}Ours \\ (enhanced, fine-tuned)\end{tabular} &  & \textbf{\resft} \\ \hline
\end{tabular}
        \end{center}
        \label{tab:resnet}
    \end{table}

    ResNet complexity uniformly drops on each residual block, as is shown in Table~\ref{tab:rescomp}.
    Guided by single layer experiments (Sec. \ref{sec:ablation}),
    we still prefer reducing shallower layers heavier than deeper ones. 
    
    Following similar setting as \filterpruning~\cite{Li2016},
    we keep 70\% channels for sensitive residual blocks (\verb|res5| and blocks close to the position where spatial size change, \eg \verb|res3a,res3d|). 
    As for other blocks, we keep 30\% channels.
    With the multi-branch enhancement, we prune \verb|branch2a| more aggressively within each residual block.
    The preserving channels ratios for \verb|branch2a,branch2b,branch2c| is $2:4:3$ (\eg, Given 30\%, we keep 40\%, 80\%, 60\% respectively).
    
    We evaluate the performance of multi-branch variants of our approach (Sec. \ref{sec:multi}).
    Our approach outperforms recent approaches (ThiNet~\cite{luo2017thinet}, Structured Probabilistic Pruning~\cite{wang2017structured}) by a large margin.
    From Table~\ref{tab:resnet}, we improve 4.0\% 
    with our multi-branch enhancement.
    This is because we accounted the accumulated error from shortcut connection which could broadcast to every layer after it.
    And the large input \featch\ at the entry of each residual block is well reduced by our \sampling.

   \subsubsection{Xception Pruning}

          \begin{table}
        \caption{Comparisons for \xceptionfifty, under \x{2} acceleration ratio. The baseline network's top-5 accuracy is \xceptionorig\%. Our approach outperforms previous approaches. Most \structured\ methods are not effective on Xception architecture (\textit{smaller is better}).}
        \begin{center}
           \begin{tabular}{|c|c|}
               \hline
               Solution                      & Increased err.        \\ \hline
               \pr                                                                     & \xceptionorig        \\ \hline
               \prfttwo &  \xceptionpr          \\ \hline
               Ours              & \xceptioncr          \\ \hline
               Ours (fine-tuned)       &\textbf{\xceptionft} \\ \hline
           \end{tabular}
       \end{center}
       \label{tab:xception}
   \end{table}    
   Since computational complexity becomes important in model design, 
   separable convolution has been payed much attention~\cite{xie2016aggregated,chollet2016xception}.
   Xception~\cite{chollet2016xception} is already spatially optimized and \tensordecom\ on $1\times1$ \conv\ is destructive.
   Thanks to our approach, it could still be accelerated with graceful degradation.
   For the ease of comparison, we adopt Xception convolution on ResNet-50, denoted by Xception-50.
   Based on ResNet-50, we swap all \conv s with spatial conv blocks.
   To keep the same computational complexity, 
   we increase the input channels of all \verb|branch2b| layers by \x{2}.
   The baseline Xception-50\footnote{https://github.com/yihui-he/Xception-caffe} has a top-5 accuracy of \xceptionorig\% and complexity of 4450 MFLOPs.
   
   We apply multi-branch variants of our approach as described in Sec. \ref{sec:multi}, 
   and adopt the same pruning ratio setting as ResNet in the previous section.
   Maybe because of Xception block is unstable,
   Batch Normalization layers must be maintained during pruning.
   Otherwise, it becomes non-trivial to fine-tune the pruned model.

   Shown in Table~\ref{tab:xception},
   after fine-tuning, we only suffer \textbf{\xceptionft\%} increase of error under \x{2}.
   \prli\ could also apply on Xception, though it is designed for small \ratio.
   Without fine-tuning, the top-5 error is 100\%. 
   After training 20 epochs, the increased error reach \xceptionpr\% which is like training from scratch.
   Our results for Xception-50 are not as graceful as results for VGG-16 since modern networks tend to have less redundancy by design.
   
       \subsubsection{Experiments on CIFAR-10}
       Even though our approach is designed for large datasets,
       it could generalize well on small datasets.
       We perform experiments on CIFAR-10 dataset~\cite{Krizhevsky2009}, which is favored by many acceleration studies.
       It consists of 50k images for training and 10k for testing in 10 classes. 
       The original $32\times32$ image is zero padded with 4 pixels on each side, 
    then random cropped to $32\times32$ at training time.
    Our approach could be finished within minutes.

   \begin{table*}
    \caption{\x{1.4} and \x{2} speed-up comparisons for ResNet-56 on CIFAR-10, the baseline accuracy is 92.8\% (one view). We outperforms previous approaches and scratch trained counterpart (\textit{smaller is better}). 
    }
    \begin{center}
\begin{tabular}{|c|c|c|c|c|}
\hline
Model & \multicolumn{2}{c|}{\x{1.4}}            & \multicolumn{2}{c|}{\x{2}} \\ \hline
                                             & err.           & increased err.         & err.    & increased err.   \\ \hline
\prft                                   & 7.2            & \textbf{0.0}           & 8.5     & 1.3              \\ \hline
From scratch                                 & 7.8            & 0.6                    & 9.1     & 1.9              \\ \hline
Ours                                         & 7.7            & 0.5                    & 9.2     & 2.0              \\ \hline
Ours (fine-tuned)                            & 7.2            & \textbf{0.0}           & 8.2     & \textbf{1.0}     \\ \hline
from scratch (uniformed)                     & \multicolumn{2}{c|}{\multirow{3}{*}{-}} & 8.7     & 1.5              \\ \cline{1-1} \cline{4-5} 
ours (uniformed)                             & \multicolumn{2}{c|}{}                   & 10.2    & 3.0              \\ \cline{1-1} \cline{4-5} 
ours (uniformed, fine-tuned)                 & \multicolumn{2}{c|}{}                   & 8.6     & \textbf{1.4}     \\ \hline
\end{tabular}
\end{center}
    \label{tab:cifar}
    \end{table*}

       We reproduce ResNet-56\footnote{https://github.com/yihui-he/resnet-cifar10-caffe}, which has accuracy of 92.8\% (Serve as a reference, the official ResNet-56~\cite{He2015} has accuracy of 93.0\%).
       For \x{2} acceleration, we follow similar setting as Sec.~\ref{sec:resnetimagenet} (keep the final stage unchanged, where the spatial size is $8\times8$).
       Shown in Table~\ref{tab:cifar}, our approach is competitive with scratch trained one, without fine-tuning,
       under both \x{1.4} and 
       \x{2} speed-up.
       After fine-tuning, our result is significantly better than \filterpruning~\cite{Li2016} and scratch trained one for both \x{1.4} and \x{2} acceleration.

      \textbf{\textit{Reduce Shallow vs. Deep layers}}:
      Denoted by \textit{(uniform)} in Table \ref{tab:cifar}, uniformed solution prune each layer while the same pruning ratio.
      Clearly, our uniformed results are worse than shallow layers heavily reduced ones.
      However, uniformed model from scratched is better than its counterpart.
      This is because \channelpruning\ favors less channels on shallow layers, however from scratch models performs better with more shallower layers.
      It indicates that redundancy on shallow layers is necessary while training, which could be removed at inference time.

\section{Conclusion}
To conclude, current deep CNNs are accurate with high inference costs.
In this paper, we have presented an inference-time \channelpruning\ method for very deep networks.
The reduced CNNs are inference efficient networks while maintaining accuracy,
and only require off-the-shelf libraries.
Compelling speed-ups and accuracy are demonstrated for both VGG Net and ResNet-like networks
on ImageNet, CIFAR-10 and PASCAL VOC 2007.

In the future, we plan to involve our approaches to training time to accelerate training procedure, instead of inference time only.

\ifCLASSOPTIONcaptionsoff
  \newpage
\fi

\bibliographystyle{IEEEtran}
\bibliography{egbib}

\begin{thebibliography}{10}
\providecommand{\url}[1]{#1}
\csname url@samestyle\endcsname
\providecommand{\newblock}{\relax}
\providecommand{\bibinfo}[2]{#2}
\providecommand{\BIBentrySTDinterwordspacing}{\spaceskip=0pt\relax}
\providecommand{\BIBentryALTinterwordstretchfactor}{4}
\providecommand{\BIBentryALTinterwordspacing}{\spaceskip=\fontdimen2\font plus
\BIBentryALTinterwordstretchfactor\fontdimen3\font minus
  \fontdimen4\font\relax}
\providecommand{\BIBforeignlanguage}[2]{{%
\expandafter\ifx\csname l@#1\endcsname\relax
\typeout{** WARNING: IEEEtran.bst: No hyphenation pattern has been}%
\typeout{** loaded for the language `#1'. Using the pattern for}%
\typeout{** the default language instead.}%
\else
\language=\csname l@#1\endcsname
\fi
#2}}
\providecommand{\BIBdecl}{\relax}
\BIBdecl

\bibitem{szegedy2015rethinking}
C.~Szegedy, V.~Vanhoucke, S.~Ioffe, J.~Shlens, and Z.~Wojna, ``Rethinking the
  inception architecture for computer vision,'' \emph{arXiv preprint
  arXiv:1512.00567}, 2015.

\bibitem{vasilache2014fast}
N.~Vasilache, J.~Johnson, M.~Mathieu, S.~Chintala, S.~Piantino, and Y.~LeCun,
  ``Fast convolutional nets with fbfft: A gpu performance evaluation,''
  \emph{arXiv preprint arXiv:1412.7580}, 2014.

\bibitem{courbariaux2016binarynet}
M.~Courbariaux and Y.~Bengio, ``Binarynet: Training deep neural networks with
  weights and activations constrained to+ 1 or-1,'' \emph{arXiv preprint
  arXiv:1602.02830}, 2016.

\bibitem{jaderberg2014speeding}
M.~Jaderberg, A.~Vedaldi, and A.~Zisserman, ``Speeding up convolutional neural
  networks with low rank expansions,'' \emph{arXiv preprint arXiv:1405.3866},
  2014.

\bibitem{han2015learning}
S.~Han, J.~Pool, J.~Tran, and W.~Dally, ``Learning both weights and connections
  for efficient neural network,'' in \emph{Advances in Neural Information
  Processing Systems}, 2015, pp. 1135--1143.

\bibitem{wen2016learning}
W.~Wen, C.~Wu, Y.~Wang, Y.~Chen, and H.~Li, ``Learning structured sparsity in
  deep neural networks,'' in \emph{Advances In Neural Information Processing
  Systems}, 2016, pp. 2074--2082.

\bibitem{szegedy2015going}
C.~Szegedy, W.~Liu, Y.~Jia, P.~Sermanet, S.~Reed, D.~Anguelov, D.~Erhan,
  V.~Vanhoucke, and A.~Rabinovich, ``Going deeper with convolutions,'' in
  \emph{Proceedings of the IEEE Conference on Computer Vision and Pattern
  Recognition}, 2015, pp. 1--9.

\bibitem{He2015}
K.~He, X.~Zhang, S.~Ren, and J.~Sun, ``Deep residual learning for image
  recognition,'' in \emph{Proceedings of the IEEE conference on computer vision
  and pattern recognition}, 2016, pp. 770--778.

\bibitem{chollet2016xception}
F.~Chollet, ``Xception: Deep learning with depthwise separable convolutions,''
  \emph{arXiv preprint arXiv:1610.02357}, 2016.

\bibitem{Alvarez2016}
J.~M. Alvarez and M.~Salzmann, ``Learning the number of neurons in deep
  networks,'' in \emph{Advances in Neural Information Processing Systems},
  2016, pp. 2262--2270.

\bibitem{Li2016}
H.~Li, A.~Kadav, I.~Durdanovic, H.~Samet, and H.~P. Graf, ``Pruning filters for
  efficient convnets,'' \emph{arXiv preprint arXiv:1608.08710}, 2016.

\bibitem{anwar2016compact}
S.~Anwar and W.~Sung, ``Compact deep convolutional neural networks with coarse
  pruning,'' \emph{arXiv preprint arXiv:1610.09639}, 2016.

\bibitem{Chen2015}
T.~Chen, I.~Goodfellow, and J.~Shlens, ``Net2net: Accelerating learning via
  knowledge transfer,'' \emph{arXiv preprint arXiv:1511.05641}, 2015.

\bibitem{Zhang2015}
X.~Zhang, J.~Zou, K.~He, and J.~Sun, ``Accelerating very deep convolutional
  networks for classification and detection,'' \emph{IEEE transactions on
  pattern analysis and machine intelligence}, vol.~38, no.~10, pp. 1943--1955,
  2016.

\bibitem{luo2017thinet}
J.-H. Luo, J.~Wu, and W.~Lin, ``Thinet: A filter level pruning method for deep
  neural network compression,'' in \emph{Proceedings of the IEEE Conference on
  Computer Vision and Pattern Recognition}, 2017, pp. 5058--5066.

\bibitem{lin2017runtime}
J.~Lin, Y.~Rao, and J.~Lu, ``Runtime neural pruning,'' in \emph{Advances in
  Neural Information Processing Systems}, 2017, pp. 2178--2188.

\bibitem{wang2017structured}
H.~Wang, Q.~Zhang, Y.~Wang, and R.~Hu, ``Structured probabilistic pruning for
  deep convolutional neural network acceleration,'' \emph{arXiv preprint
  arXiv:1709.06994}, 2017.

\bibitem{liu2017learning}
Z.~Liu, J.~Li, Z.~Shen, G.~Huang, S.~Yan, and C.~Zhang, ``Learning efficient
  convolutional networks through network slimming,'' in \emph{Proceedings of
  the IEEE Conference on Computer Vision and Pattern Recognition}, 2017, pp.
  2736--2744.

\bibitem{He_2017_ICCV}
Y.~He, X.~Zhang, and J.~Sun, ``Channel pruning for accelerating very deep
  neural networks,'' in \emph{The IEEE International Conference on Computer
  Vision (ICCV)}, Oct 2017, pp. 1389--1397.

\bibitem{cheng2017survey}
Y.~Cheng, D.~Wang, P.~Zhou, and T.~Zhang, ``A survey of model compression and
  acceleration for deep neural networks,'' \emph{arXiv preprint
  arXiv:1710.09282}, 2017.

\bibitem{lecun1989optimal}
Y.~LeCun, J.~S. Denker, S.~A. Solla, R.~E. Howard, and L.~D. Jackel, ``Optimal
  brain damage.'' in \emph{NIPs}, vol.~2, 1989, pp. 598--605.

\bibitem{hassibi1993second}
B.~Hassibi and D.~G. Stork, \emph{Second order derivatives for network pruning:
  Optimal brain surgeon}.\hskip 1em plus 0.5em minus 0.4em\relax Morgan
  Kaufmann, 1993.

\bibitem{bagherinezhad2016lcnn}
H.~Bagherinezhad, M.~Rastegari, and A.~Farhadi, ``Lcnn: Lookup-based
  convolutional neural network,'' \emph{arXiv preprint arXiv:1611.06473}, 2016.

\bibitem{Rastegari2016}
M.~Rastegari, V.~Ordonez, J.~Redmon, and A.~Farhadi, ``Xnor-net: Imagenet
  classification using binary convolutional neural networks,'' in
  \emph{European Conference on Computer Vision}.\hskip 1em plus 0.5em minus
  0.4em\relax Springer, 2016, pp. 525--542.

\bibitem{mathieu2013fast}
M.~Mathieu, M.~Henaff, and Y.~LeCun, ``Fast training of convolutional networks
  through ffts,'' \emph{arXiv preprint arXiv:1312.5851}, 2013.

\bibitem{lavin2015fast}
A.~Lavin, ``Fast algorithms for convolutional neural networks,'' \emph{arXiv
  preprint arXiv:1509.09308}, 2015.

\bibitem{phan2020mobinet}
H.~Phan, Y.~He, M.~Savvides, Z.~Shen \emph{et~al.}, ``Mobinet: A mobile binary
  network for image classification,'' in \emph{Proceedings of the IEEE/CVF
  Winter Conference on Applications of Computer Vision}, 2020, pp. 3453--3462.

\bibitem{liu2015sparse}
B.~Liu, M.~Wang, H.~Foroosh, M.~Tappen, and M.~Pensky, ``Sparse convolutional
  neural networks,'' in \emph{Proceedings of the IEEE Conference on Computer
  Vision and Pattern Recognition}, 2015, pp. 806--814.

\bibitem{lebedev2015fast}
V.~Lebedev and V.~Lempitsky, ``Fast convnets using group-wise brain damage,''
  \emph{arXiv preprint arXiv:1506.02515}, 2015.

\bibitem{han2016eie}
S.~Han, X.~Liu, H.~Mao, J.~Pu, A.~Pedram, M.~A. Horowitz, and W.~J. Dally,
  ``Eie: efficient inference engine on compressed deep neural network,'' in
  \emph{Proceedings of the 43rd International Symposium on Computer
  Architecture}.\hskip 1em plus 0.5em minus 0.4em\relax IEEE Press, 2016, pp.
  243--254.

\bibitem{guo2016dynamic}
Y.~Guo, A.~Yao, and Y.~Chen, ``Dynamic network surgery for efficient dnns,'' in
  \emph{Advances In Neural Information Processing Systems}, 2016, pp.
  1379--1387.

\bibitem{zhong2018shift}
H.~Zhong, X.~Liu, Y.~He, and Y.~Ma, ``Shift-based primitives for efficient
  convolutional neural networks,'' \emph{arXiv preprint arXiv:1809.08458},
  2018.

\bibitem{he2019addressnet}
Y.~He, X.~Liu, H.~Zhong, and Y.~Ma, ``Addressnet: Shift-based primitives for
  efficient convolutional neural networks,'' in \emph{2019 IEEE Winter
  conference on applications of computer vision (WACV)}.\hskip 1em plus 0.5em
  minus 0.4em\relax IEEE, 2019, pp. 1213--1222.

\bibitem{yang2016designing}
T.-J. Yang, Y.-H. Chen, and V.~Sze, ``Designing energy-efficient convolutional
  neural networks using energy-aware pruning,'' \emph{arXiv preprint
  arXiv:1611.05128}, 2016.

\bibitem{han2015deep}
S.~Han, H.~Mao, and W.~J. Dally, ``Deep compression: Compressing deep neural
  network with pruning, trained quantization and huffman coding,'' \emph{CoRR,
  abs/1510.00149}, vol.~2, 2015.

\bibitem{lebedev2014speeding}
V.~Lebedev, Y.~Ganin, M.~Rakhuba, I.~Oseledets, and V.~Lempitsky, ``Speeding-up
  convolutional neural networks using fine-tuned cp-decomposition,''
  \emph{arXiv preprint arXiv:1412.6553}, 2014.

\bibitem{gong2014compressing}
Y.~Gong, L.~Liu, M.~Yang, and L.~Bourdev, ``Compressing deep convolutional
  networks using vector quantization,'' \emph{arXiv preprint arXiv:1412.6115},
  2014.

\bibitem{kim2015compression}
Y.-D. Kim, E.~Park, S.~Yoo, T.~Choi, L.~Yang, and D.~Shin, ``Compression of
  deep convolutional neural networks for fast and low power mobile
  applications,'' \emph{arXiv preprint arXiv:1511.06530}, 2015.

\bibitem{he2019depth}
Y.~He, J.~Qian, and J.~Wang, ``Depth-wise decomposition for accelerating
  separable convolutions in efficient convolutional neural networks,'' in
  \emph{Proceedings of the IEEE Conference on Computer Vision and Pattern
  Recognition Workshops}, 2019.

\bibitem{xue2013restructuring}
J.~Xue, J.~Li, and Y.~Gong, ``Restructuring of deep neural network acoustic
  models with singular value decomposition.'' in \emph{INTERSPEECH}, 2013, pp.
  2365--2369.

\bibitem{denton2014exploiting}
E.~L. Denton, W.~Zaremba, J.~Bruna, Y.~LeCun, and R.~Fergus, ``Exploiting
  linear structure within convolutional networks for efficient evaluation,'' in
  \emph{Advances in Neural Information Processing Systems}, 2014, pp.
  1269--1277.

\bibitem{girshick2015fast}
R.~Girshick, ``Fast r-cnn,'' in \emph{Proceedings of the IEEE International
  Conference on Computer Vision}, 2015, pp. 1440--1448.

\bibitem{zhang2020image}
X.~Zhang and H.~Yihui, ``Image processing method and apparatus, and
  computer-readable storage medium,'' Jul.~14 2020, uS Patent 10,713,533.

\bibitem{zhou2016less}
H.~Zhou, J.~M. Alvarez, and F.~Porikli, ``Less is more: Towards compact cnns,''
  in \emph{European Conference on Computer Vision}.\hskip 1em plus 0.5em minus
  0.4em\relax Springer International Publishing, 2016, pp. 662--677.

\bibitem{lecun1998gradient}
Y.~LeCun, L.~Bottou, Y.~Bengio, and P.~Haffner, ``Gradient-based learning
  applied to document recognition,'' \emph{Proceedings of the IEEE}, vol.~86,
  no.~11, pp. 2278--2324, 1998.

\bibitem{krizhevsky2012imagenet}
A.~Krizhevsky, I.~Sutskever, and G.~E. Hinton, ``Imagenet classification with
  deep convolutional neural networks,'' in \emph{Advances in neural information
  processing systems}, 2012, pp. 1097--1105.

\bibitem{anwar2015structured}
S.~Anwar, K.~Hwang, and W.~Sung, ``Structured pruning of deep convolutional
  neural networks,'' \emph{arXiv preprint arXiv:1512.08571}, 2015.

\bibitem{polyak2015channel}
A.~Polyak and L.~Wolf, ``Channel-level acceleration of deep face
  representations,'' \emph{IEEE Access}, vol.~3, pp. 2163--2175, 2015.

\bibitem{He_2018_ECCV}
Y.~He, J.~Lin, Z.~Liu, H.~Wang, L.-J. Li, and S.~Han, ``Amc: Automl for model
  compression and acceleration on mobile devices,'' in \emph{European
  Conference on Computer Vision}, Sept 2018.

\bibitem{srinivas2015data}
S.~Srinivas and R.~V. Babu, ``Data-free parameter pruning for deep neural
  networks,'' \emph{arXiv preprint arXiv:1507.06149}, 2015.

\bibitem{mariet2015diversity}
Z.~Mariet and S.~Sra, ``Diversity networks,'' \emph{arXiv preprint
  arXiv:1511.05077}, 2015.

\bibitem{hu2016network}
H.~Hu, R.~Peng, Y.-W. Tai, and C.-K. Tang, ``Network trimming: A data-driven
  neuron pruning approach towards efficient deep architectures,'' \emph{arXiv
  preprint arXiv:1607.03250}, 2016.

\bibitem{tibshirani1996regression}
R.~Tibshirani, ``Regression shrinkage and selection via the lasso,''
  \emph{Journal of the Royal Statistical Society. Series B (Methodological)},
  pp. 267--288, 1996.

\bibitem{breiman1995better}
L.~Breiman, ``Better subset regression using the nonnegative garrote,''
  \emph{Technometrics}, vol.~37, no.~4, pp. 373--384, 1995.

\bibitem{Simonyan2014}
K.~Simonyan and A.~Zisserman, ``Very deep convolutional networks for
  large-scale image recognition,'' \emph{arXiv preprint arXiv:1409.1556}, 2014.

\bibitem{JiaDeng2009}
J.~Deng, W.~Dong, R.~Socher, L.-J. Li, K.~Li, and L.~Fei-Fei, ``Imagenet: A
  large-scale hierarchical image database,'' in \emph{Computer Vision and
  Pattern Recognition, 2009. CVPR 2009. IEEE Conference on}.\hskip 1em plus
  0.5em minus 0.4em\relax IEEE, 2009, pp. 248--255.

\bibitem{Krizhevsky2009}
A.~Krizhevsky and G.~Hinton, ``Learning multiple layers of features from tiny
  images,'' 2009.

\bibitem{pascal-voc-2007}
M.~Everingham, L.~Van~Gool, C.~K.~I. Williams, J.~Winn, and A.~Zisserman, ``The
  {PASCAL} {V}isual {O}bject {C}lasses {C}hallenge 2007 {(VOC2007)}
  {R}esults,''
  http://www.pascal-network.org/challenges/VOC/voc2007/workshop/index.html.

\bibitem{ioffe2015batch}
S.~Ioffe and C.~Szegedy, ``Batch normalization: Accelerating deep network
  training by reducing internal covariate shift,'' \emph{arXiv preprint
  arXiv:1502.03167}, 2015.

\bibitem{nair2010rectified}
V.~Nair and G.~E. Hinton, ``Rectified linear units improve restricted boltzmann
  machines,'' in \emph{Proceedings of the 27th international conference on
  machine learning (ICML-10)}, 2010, pp. 807--814.

\bibitem{jia2014caffe}
Y.~Jia, E.~Shelhamer, J.~Donahue, S.~Karayev, J.~Long, R.~Girshick,
  S.~Guadarrama, and T.~Darrell, ``Caffe: Convolutional architecture for fast
  feature embedding,'' \emph{arXiv preprint arXiv:1408.5093}, 2014.

\bibitem{tensorflow}
\BIBentryALTinterwordspacing
M.~Abadi, A.~Agarwal, P.~Barham, E.~Brevdo, Z.~Chen, C.~Citro, G.~S. Corrado,
  A.~Davis, J.~Dean, M.~Devin, S.~Ghemawat, I.~Goodfellow, A.~Harp, G.~Irving,
  M.~Isard, Y.~Jia, R.~Jozefowicz, L.~Kaiser, M.~Kudlur, J.~Levenberg,
  D.~Man\'{e}, R.~Monga, S.~Moore, D.~Murray, C.~Olah, M.~Schuster, J.~Shlens,
  B.~Steiner, I.~Sutskever, K.~Talwar, P.~Tucker, V.~Vanhoucke, V.~Vasudevan,
  F.~Vi\'{e}gas, O.~Vinyals, P.~Warden, M.~Wattenberg, M.~Wicke, Y.~Yu, and
  X.~Zheng, ``{TensorFlow}: Large-scale machine learning on heterogeneous
  systems,'' 2015, software available from tensorflow.org. [Online]. Available:
  \url{http://tensorflow.org/}
\BIBentrySTDinterwordspacing

\bibitem{scikit-learn}
F.~Pedregosa, G.~Varoquaux, A.~Gramfort, V.~Michel, B.~Thirion, O.~Grisel,
  M.~Blondel, P.~Prettenhofer, R.~Weiss, V.~Dubourg, J.~Vanderplas, A.~Passos,
  D.~Cournapeau, M.~Brucher, M.~Perrot, and E.~Duchesnay, ``Scikit-learn:
  Machine learning in {P}ython,'' \emph{Journal of Machine Learning Research},
  vol.~12, pp. 2825--2830, 2011.

\bibitem{huang2016speed}
J.~Huang, V.~Rathod, C.~Sun, M.~Zhu, A.~Korattikara, A.~Fathi, I.~Fischer,
  Z.~Wojna, Y.~Song, S.~Guadarrama \emph{et~al.}, ``Speed/accuracy trade-offs
  for modern convolutional object detectors,'' \emph{arXiv preprint
  arXiv:1611.10012}, 2016.

\bibitem{cuda}
\BIBentryALTinterwordspacing
J.~Nickolls, I.~Buck, M.~Garland, and K.~Skadron, ``Scalable parallel
  programming with {CUDA},'' \emph{{ACM} Queue}, vol.~6, no.~2, pp. 40--53,
  2008. [Online]. Available: \url{http://doi.acm.org/10.1145/1365490.1365500}
\BIBentrySTDinterwordspacing

\bibitem{cudnn}
\BIBentryALTinterwordspacing
S.~Chetlur, C.~Woolley, P.~Vandermersch, J.~Cohen, J.~Tran, B.~Catanzaro, and
  E.~Shelhamer, ``cudnn: Efficient primitives for deep learning,'' \emph{CoRR},
  vol. abs/1410.0759, 2014. [Online]. Available:
  \url{http://arxiv.org/abs/1410.0759}
\BIBentrySTDinterwordspacing

\bibitem{yolo}
\BIBentryALTinterwordspacing
J.~Redmon, S.~K. Divvala, R.~B. Girshick, and A.~Farhadi, ``You only look once:
  Unified, real-time object detection,'' \emph{CoRR}, vol. abs/1506.02640,
  2015. [Online]. Available: \url{http://arxiv.org/abs/1506.02640}
\BIBentrySTDinterwordspacing

\bibitem{ssd}
\BIBentryALTinterwordspacing
W.~Liu, D.~Anguelov, D.~Erhan, C.~Szegedy, S.~E. Reed, C.~Fu, and A.~C. Berg,
  ``{SSD:} single shot multibox detector,'' \emph{CoRR}, vol. abs/1512.02325,
  2015. [Online]. Available: \url{http://arxiv.org/abs/1512.02325}
\BIBentrySTDinterwordspacing

\bibitem{zhu2019feature}
C.~Zhu, Y.~He, and M.~Savvides, ``Feature selective anchor-free module for
  single-shot object detection,'' in \emph{Proceedings of the IEEE/CVF
  conference on computer vision and pattern recognition}, 2019, pp. 840--849.

\bibitem{he2020deep}
Y.~He and J.~Wang, ``Deep mixture density network for probabilistic object
  detection,'' in \emph{2020 IEEE/RSJ International Conference on Intelligent
  Robots and Systems (IROS)}.\hskip 1em plus 0.5em minus 0.4em\relax IEEE,
  2020, pp. 10\,550--10\,555.

\bibitem{he2019bounding}
Y.~He, C.~Zhu, J.~Wang, M.~Savvides, and X.~Zhang, ``Bounding box regression
  with uncertainty for accurate object detection,'' in \emph{Proceedings of the
  ieee/cvf conference on computer vision and pattern recognition}, 2019, pp.
  2888--2897.

\bibitem{he2019deep}
Y.~He and J.~Wang, ``Deep multivariate mixture of gaussians for object
  detection under occlusion,'' 2019.

\bibitem{he2018softer}
Y.~He, X.~Zhang, M.~Savvides, and K.~Kitani, ``Softer-nms: Rethinking bounding
  box regression for accurate object detection,'' \emph{arXiv preprint
  arXiv:1809.08545}, vol.~2, no.~3, pp. 69--80, 2018.

\bibitem{fasterrcnn}
\BIBentryALTinterwordspacing
S.~Ren, K.~He, R.~B. Girshick, and J.~Sun, ``Faster {R-CNN:} towards real-time
  object detection with region proposal networks,'' \emph{CoRR}, vol.
  abs/1506.01497, 2015. [Online]. Available:
  \url{http://arxiv.org/abs/1506.01497}
\BIBentrySTDinterwordspacing

\bibitem{lin2014microsoft}
T.-Y. Lin, M.~Maire, S.~Belongie, J.~Hays, P.~Perona, D.~Ramanan,
  P.~Doll{\'a}r, and C.~L. Zitnick, ``Microsoft coco: Common objects in
  context,'' in \emph{European conference on computer vision}.\hskip 1em plus
  0.5em minus 0.4em\relax Springer, Cham, 2014, pp. 740--755.

\bibitem{xie2016aggregated}
S.~Xie, R.~Girshick, P.~Doll{\'a}r, Z.~Tu, and K.~He, ``Aggregated residual
  transformations for deep neural networks,'' \emph{arXiv preprint
  arXiv:1611.05431}, 2016.

\end{thebibliography}

\vfill

\end{document}